%% file: main.tex
\pdfoutput=1

\documentclass[11pt]{article}

\usepackage[preprint]{acl}

\usepackage{newtxmath}
\usepackage{newtxtext}

\usepackage{fix-cm}
\usepackage{latexsym}

\usepackage[T1]{fontenc}

\usepackage[utf8]{inputenc}

\usepackage{microtype}

\usepackage{inconsolata}

\usepackage{graphicx}

\usepackage{booktabs}
\usepackage{standalone}
\usepackage{tikz}
\usepackage{pgfplots}
\usepackage{siunitx}
\usepackage{multirow, xcolor, colortbl}
\usepackage{soul}
\usepackage{hhline}
\usepackage{etoolbox}
\usepackage[capitalize]{cleveref}
\usepackage{caption}
\usepackage{subcaption}
\usepackage{relsize}
\usepackage{threeparttable}
\usepackage{tabularx}

\pgfplotsset{compat=newest}
\usepgfplotslibrary{groupplots}

\robustify\bfseries
\usetikzlibrary{positioning, patterns, fit}

\usepackage{xspace}
\newcommand\ours{SD-RA-IT\xspace}
\newcommand\baseline{RA-IT\xspace}
\newcommand\llamainstruct{Llama-3-Instruct\xspace}
\newcommand\llamasmall{Llama-3-8B-Instruct\xspace}
\newcommand\llamabig{Llama-3-70B-Instruct\xspace}
\newcommand\llamahuge{Llama-3.1-405B-Instruct\xspace}
\title{Post-training an LLM for RAG? Train on Self-Generated Demonstrations}

\author{
  Matthew Finlayson\footnotemark[1]
  \quad Ilia Kulikov\footnotemark[2]
  \quad Daniel M. Bikel\footnotemark[2] \\
  \bf \quad Barlas Oguz\footnotemark[2]
  \quad Xilun Chen\footnotemark[2] 
  \quad Aasish Pappu\footnotemark[2] \\
  \footnotemark[1]University of Southern California
  \quad \footnotemark[2]Meta, Inc.
  \\
  \texttt{mfinlays@usc.edu}\quad\texttt{aasish@meta.com}
}

\begin{document}
\maketitle

\begin{abstract}
  \input{abstract}
\end{abstract}

\begin{figure*}
  \centering
  \small
  \input{fig/baseline}
  \caption[Placeholder caption]{
    An overview of existing of na\"ive retrieval-augmented instruction tuning 
    (training on OOD data) 
    and our proposed method: 
    retrieval-augmented instruction tuning on self-generated demos.
    Fine-tuning on out-of-distribution (OOD) instruction responses (highlighted \sethlcolor{pink}\hl{red}) from an outside source,
    for example an existing annotated dataset, can degrade the model.
    We use a model to generate its own correct, in-distribution responses (highlighted \sethlcolor{green}\hl{green}). 
    Training on these yields better performance on downstream RAG tasks.
  }
  \label{fig1}
\end{figure*}
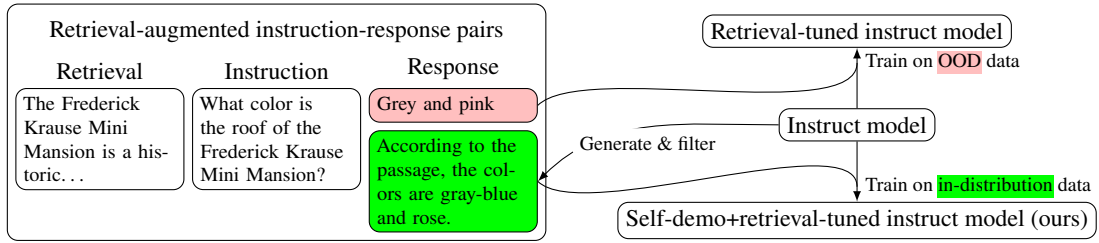

\section{Introduction}

Typically, large language models (LLMs) must rely solely on knowledge learned during pre-training to perform downstream tasks like question answering.
This situation becomes problematic when the task depends on knowledge that the model has not encountered or learned.
A popular framework for mitigating this issue is to provide the model with \textit{contextual knowledge}, i.e., providing relevant information for the task as input.
This crucial context typically comes from \emph{retrieval systems} that fetch documents from some database based on their relevance to the task.
The combination of retrieval and language generation,
known as retrieval-augmented generation (RAG),
often outperforms LLMs without retrievals on downstream tasks,
particularly for knowledge intensive tasks where parametric (i.e., learned during pre-training) knowledge fails.

Integrating an LLM into a RAG system can be tricky,
since the LLM may not properly leverage the additional context to improve performance.
In particular, if the retrievals are irrelevant to the task, they may distract the LLM from the objective or cause the LLM to integrate incorrect information into its generations.
The LLM may also fail to integrate the provided context even when it contains the solution to the task.
We hypothesize that these shortcomings are caused by distributional mismatch between the model's training data and the RAG-style inputs it receives at test time.

Several prior efforts have sought to better integrate LLMs into RAG systems 
by fine-tuning them for the task.
This usually involves training the model with in-context retrievals,
and may be implemented during pre-training~\citep[e.g.,][]{guu2020retrieval},
or as a post-training step~\citep{lin2024radit}.
Since pre-training can be prohibitively expensive,
we focus on the more accessible and common post-training methods.
In particular, we concentrate on retrieval-augmented instruction tuning (\baseline) 
where an LLM is fine-tuned on an instruction tuning dataset which has been modified to include one or more retrievals for each instance.

One easy way to obtain a \baseline training set is to augment existing instruction-response pairs from an instruction tuning dataset with retrievals. This approach poses two significant challenges.
First, there may be a misalignment between the retrieved content and the response, as the latter was created without reference to the retrieval and might even appear to contradict it. 
Second---and this is a more general issue in fine-tuning---gold responses are often unlikely to be generated by the model, making them out-of-distribution (OOD).
Training on OOD responses can inadvertently promote hallucination by compelling the model to generate outputs that conflict with its parametric knowledge~\citep{Lin2024FLAMEFA}.
Consequently, \baseline may elicit undesirable LLM behavior.

We propose to remedy the OOD response issue by replacing responses in a \baseline dataset
with ones that match the capabilities and distribution of the LLM. 
We accomplish this by obtaining \emph{self-generated demonstrations} (self-demos) from the model, as illustrated in \cref{fig1}.
Our process consists of generating multiple outputs from the LLM conditioned on retrievals and instructions, then filtering for the outputs that match the gold responses (as judged by an LLM).
Our generate-then-filter process retains ground truth supervision while matching the training data to the model distribution.
To maximize the number of successful self-demos, we sample from multiple prompts and generate both with and without retrievals.
If none of the response candidates match the gold response we instead select a refusal generated based on the context.
Thus, we obtain self-demos that match the gold responses, are in-distribution for the LLM, and respect its parametric and contextual knowledge.

We find that models obtained via self-demo retrieval-augmented instruction tuning (\ours) perform favorably on knowledge-intensive question answering tasks compared to other methods.
Their answer attempts succeed at a higher rate, 
and they better identify answers from retrievals.
They accomplish this by abstaining from questions they will likely get wrong,
and engage with retrievals to answer questions correctly.
We also find that, while \baseline tends to degrade the LLM's performance when retrievals are not present (non-RAG settings), \ours improves model performance across both non-RAG and many-retrieval settings, even improving performance on contexts with many more retrievals than the model was explicitly trained to handle.

\section{Related work}

Our work builds upon several areas in RAG methods and post-training.

\subsection{Training RAG-enabled LLMs}
As retrieval-augmented generation (RAG) has garnered interest as an effective way to improve LLM performance~\citep{Lewis2020RetrievalAugmentedGF}.
We focus on the setting where we prepend retrievals to the prompt, which is a common, simple method for incorporating retrievals into the context, 
though alternative RAG architectures exist~\citep[e.g.,][]{shi2023replug}. 
Our work builds upon previous studies which investigate how to train LLMs to better handle retrieved context.
Some of these propose pre-training methods that boost RAG capabilities
\citep{guu2020retrieval, Izacard2023AtlasFL, Shi2023InContextPL},
while others fine-tune a LLM for better RAG capabilities~\citet{luo-etal-2023-search}.
Our method targets the latter fine-tuning setting, where we hypothesize that out-of-distribution training examples cause performance degradation. 
Our key insight is to self-generate demonstrations, 
rather than using a teacher LLM~\citep{luo-etal-2023-search}
or human-written responses~\citep{lin2024radit}.

\subsection{Distilling prompted behavior}
Our methodology has similarities with recent work on distilling the behavior of prompted language models back into the unprompted model.
Note that this is different from distilling a larger model into a smaller one; instead, the prompted model and the target model for fine-tuning are the same.
\citet{Yu2024DistillingS2} use prompting frameworks~\citep[e.g., Chain-of-Thought, System 2 Attention;][]{wei2022chain, weston20232attentionisneed}
to generate training examples using a language model, 
then do supervised fine-tuning (SFT) to distill the behavior under these prompts back into that same model.
The resulting model exhibits behavior like the prompted models without requiring prompts. 
Our work takes a similar approach for a similar goal,
though we use automatically optimized prompts instead of hand-crafted prompting frameworks to generate training data,
and we aim to specifically distill RAG capabilities rather than reasoning abilities.
Our method, similar to the one in \citet{Yu2024DistillingS2},
also requires filtering training examples for correctness.

\subsection{Self-demo training \& learning to abstain}
Our work is partly motivated by the hypothesis \citet{Lin2024FLAMEFA} and \citeposs{kang2024unfamiliar} hypothesis that out-of-distribution training data degrade LLM during post-training and encourage hallucinations.
Similar to our work, \citet{Lin2024FLAMEFA} train LLMs on self-generated demonstrations, but with the goal of improving factuality.
They also investigate using RAG as part of their training pipeline but not during inference.
We also build on their work by introducing a filtering stage to ensure the quality and correctness of the self-generated instances before training on them. 

A large body of existing literature investigates learning when to abstain from answering questions~\citep{wen2024know}.
Our method for selecting self-demos resembles the simple method proposed by \citet{yang2023alignment}
where training examples that the model answers incorrectly are replaced with refusals. 
We incorporate and improve on their template-based refusal strategy by selecting self-generated refusals when none of the self-generated demonstrations are correct.
By using self-generated refusals instead of template-based ones, we better align the training distribution with the LLM's own distribution.

\begin{figure*}
  \centering
  \small
  \input{fig/method}
  \caption[Our method]{
    Our method uses the reference model to generate response candidates for the retrieval-instruction pairs.
    We then use the model along with the gold responses to filter the candidate responses for correct ones.
    Finally, we replace the \sethlcolor{pink}\hl{OOD} responses in the training set 
    with these \sethlcolor{green}\hl{in-distribution} self-generated responses (self-demos) and train on these. 
  }
  \label{fig:method}
\end{figure*}
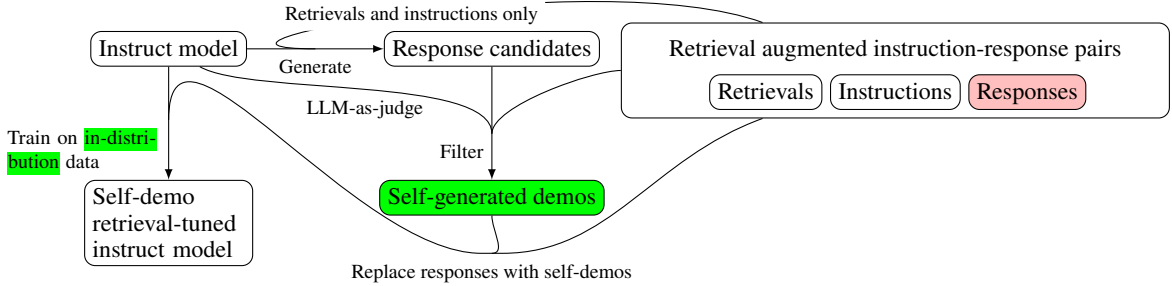

\section{Method and Implementation}
\label{sec:method}

The goal of our method (outlined in \cref{fig:method}) is to take an instruction-tuned LLM as a reference model and equip it with improved RAG capabilities.
Our method relies on the fact that instruction-tuned LLMs are already capable of several tasks,
including self-prompting, self-evaluation, question answering, and some RAG abilities.
We leverage these abilities to bootstrap a dataset of self-demonstrations to train on.

We break the process of generating self-demos for \ours into discrete steps.
First, we obtain an instruction tuning dataset augmented with retrievals.
Second, we automatically generate a set of prompts for the LLM.
Third, we generate three types of outputs:
retrieval-augmented responses,
responses without retrieval responses,
and refusals.
Fourth, we filter these responses for correctness with respect to the gold response, choosing the self-demo that is most correct, or a refusal when no outputs match the gold response.
In this section we review of our implementation of these steps.

\subsection{Models, infrastructure, and datasets}
As our reference language models, we use state-of-the-art, open-weight, instruction-tuned models \llamasmall and \llamabig.
Our method is designed to require only on the reference model; 
no larger or more powerful model is required.
We use the open-source libraries fairseq2 for training~\citep{balioglu2023fairseq2},
and vLLM for inference~\citep{kwon2023efficient},
and publicly release our training and inference scripts\footnotemark.

\makeatletter
\ifacl@anonymize
\footnotetext{Link not provided for anonymity.}
\else
\footnotetext{\url{https://github.com/facebookresearch/RAM/tree/main/projects/sd-ra-it}}
\fi
\makeatother

For convenience, we opt to use the retrieval-augmented instruction tuning dataset from \citet{lin2024radit},
who describe the composition of their dataset in Table~1 of their paper.
This dataset consists of instances which each contain an instruction, a human-written response (gold demonstration), and retrievals from Wikipedia~\citep{Izacard2023AtlasFL} and CommonCrawl\footnote{\url{https://commoncrawl.org}}.

The instruction-response pairs come from multiple domains: dialogue, open-domain QA, reading comprehension, summarization, and reasoning.
We use a mix of 10\% dialogue instances and 90\% randomly instances from other domains, following a similar split to that used in \citet{lin2024radit}.

For retrievals, we re-use the retrieved documents from \citet{lin2024radit}
which are retrieved using a DRAGON+
retriever~\citep{lin2023traindragondiverseaugmentation}.

\subsection{Automatic prompt generation}

We sample response candidates from multiple prompts with the aim to achieve diversity and maximize the likelihood of correctness.
We use automatic prompt optimization instead of hand-crafting prompts,
saving time and improving their effectiveness.
Prior research and existing libraries offer methods for automatically optimizing LLM prompts~\citep[e.g.,][]{khattab2022demonstrate, yuksekgonul2024textgrad, Reid2023AutomaticPO}\footnotemark.
Due to the simplicity of our needs, we opt to implement a custom optimizer
which roughly follows the iterative beam search formula found in the literature:
we begin with a human-written system message then (1) use the system message and reference model to generate responses to a small training set of instructions, (2) use the model to score its own outputs on a scale 1-5 compared to the gold responses, (3) use the model to critique its own outputs then suggest a new system message, sampling several new candidate system messages, then finally (4) repeat the process with the new system message candidates, pruning the lowest-scoring prompts.
\footnotetext{See \citet{AwesomeLLMPromptOptimization} for a comprehensive list of prompt optimization literature.}

After a number of rounds, we use a validation set to obtain scores for all the system message candidates, then select the top-scoring system messages.
We repeat this process for both retrieval-augmented instructions and standalone instructions.
We additionally include a human-written system message instructing the model to refuse to all instructions.

\subsection{Generating, filtering, and training on self-demonstrations}

To obtain a self-demo for each instance in our training set,
we first generate multiple output candidates using the system prompts collected in the previous step.
We then use the reference model as a judge, giving it access to the gold response and prompting it to select the most correct output, or to select a refusal if none of the candidates are correct.
In practice, we find that the judge struggles to choose between many candidates at once, so we select the best candidate via a single-elimination tournament.

We try two common post-training methods for LLMs.
The first, supervised instruction fine-tuning (SFT)
minimizes cross-entropy loss on the response tokens.
The second, direct preference optimization (DPO)
takes two candidate responses, a preferred response and a rejected response, and optimizes the model to maximize the probability of the preferred response while minimizing the probability of the rejected response.
We use a randomly selected self-demo candidate that was rejected during the self-demo generation process as the rejected response.

\section{Experimental details}

In this section we detail our methods and specific implementation choices. 

\subsection{Benchmarks and Baselines}

Our evaluation focuses on knowledge-intensive question-answering (QA) tasks.
For this, we use the evaluation datasets from~\citet{lin2024radit}\footnotemark,
since these datasets are popular for RAG evaluation.
\citet{lin2024radit}
provide multiple retrievals
for each of the evaluation instances. 
Each instance in these datasets consist of a question (e.g., ``who won the 2012 World Series'') and an answer (e.g., ``San Francisco Giants'').
\footnotetext{Our evaluation datasets include the Pushshift.io Reddit dataset, a previously existing dataset extracted and obtained by a third party that contains preprocessed comments posted on the social network Reddit and hosted by Pushshift.io.} 

We contextualize our model's performance with several baselines.
First, we compare to the reference model from which our model was fine-tuned.
Since the reference model is a fully-fledged instruction-tuned LLM with some pre-existing RAG capabilities, this is a strong baseline.
Next, we instruction-tune (IT) the reference model without using any retrievals.
This baseline is meant to control for the effect continued fine-tuning on our model's performance, and also to demonstrate the adverse effects of fine-tuning on OOD demonstrations.
The third baseline, most similar to our method, is retrieval-augmented instruction-tuning (\baseline) on gold demonstrations.
Both of our trained baselines (instruction tuning, \baseline) use cross-entropy loss as their training objective.
We also experimented with using DPO for \baseline, using the model's own generations as the rejected response, but found in preliminary studies that this method catastrophically degrades model accuracy on all QA tasks, likely because the gold responses are too far outside the model's distribution. 

\begin{table}
  \centering
  \small
  \begin{threeparttable}
  \begin{tabularx}{\linewidth}{lX}
    \toprule
    Acronym & Method \\
    \midrule
    IT & Instruction tuning \\
    \baseline & Retrieval-augmented instruction tuning \\
    \ours & Retrieval-augmented instruction tuning on self-demos (ours) \\
    +DPO & DPO objective used instead of SFT \\
    \midrule
    & Dataset \\
    \midrule
    PSR & Pushshift.io Reddit~\citep{fan-etal-2019-eli5} \\
    FEVER & \citet{thorne-etal-2018-fever} \\
    HPQA & HotPotQA~\citep{yang-etal-2018-hotpotqa} \\
    MMLU & \citet{hendrycks2021measuring} \\
    NQ & \citet{kwiatkowski-etal-2019-natural} \\
    TQA & Trivia QA~\citep{joshi-etal-2017-triviaqa} \\
    T-REx & \citet{elsahar-etal-2018-rex} \\
    WoW & Wizard of Wikipedia~\citep{dinan2018wizard} \\
    zsRE & \citet{levy-etal-2017-zero} \\
    \bottomrule
  \end{tabularx}
  \caption{A summary of method and dataset acronyms.}
  \end{threeparttable}
\end{table}

\subsection{Metrics} 

We aim to measure several properties of RAG models to holistically evaluate our proposed method.
In particular, we hypothesize that our training will make LLMs more accurate on questions it chooses to answer, better at learning which questions to refuse, better at integrating knowledge from retrievals, and suffer less degradation from training on OOD demonstrations. 

To measure these effects, we first label all test instances as one of \textit{correct} (if the model output matches the gold answer), \textit{refused} (if the model declines to answer), or \textit{incorrect}. 
We also label the retrievals as \textit{relevant} or \textit{irrelevant}, 
depending on whether the retrieval contains the answer to the question.
We use an LLM-as-a-judge to label our test instances rather than rely on more traditional metrics like exact match.
We do this because traditional metrics do not capture more nuanced properties of text like semantic equivalence and refusals. 
We observe that our judge LLMs are highly reliable and accurate for straightforward tasks such as determining whether model generations match the gold labels.

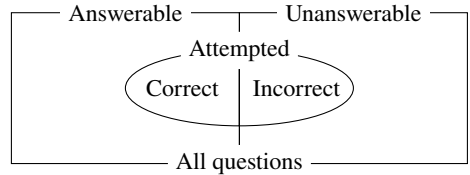
\begin{figure}
  \small
  \centering
  \begin{tikzpicture}[yscale=0.5, xscale=1.5]
    \draw (0,0) rectangle (2, 4) ;
    \node[fill=white] at (1, 4) {Answerable};
    \draw (2,0) rectangle (4, 4) ;
    \node[fill=white] at (3, 4) {Unanswerable} ;
    \draw (2, 1) arc[start angle=-90, end angle=90, radius=1cm] -- (2, 3) -- cycle;
    \node[fill=white, rounded corners] at (2.5, 2) {Incorrect} ;
    \draw (2, 3) arc[start angle=90, end angle=270, radius=1cm] -- (2, 1) -- cycle;
    \node[fill=white] at (1.5, 2) {Correct} ;
    \node[fill=white, rounded corners] at (2, 3) {Attempted} ;
    \node[fill=white, rounded corners] at (2, 0) {All questions} ;
  \end{tikzpicture}
  \caption{
    A categorization of test instances.
    The box is divided into questions that the LLM can and cannot answer correctly,
    while the oval represents questions that the LLM attempts.
    Since we are more interested in reliability than raw accuracy on downstream tasks,
    we use precision (\(\text{\# correct}/\text{\# attempted}\))
    and recall (\(\text{\# correct}/\text{\# answerable}\)) as our main metrics.
  }
  \label{fig:diagnostic_testing}
\end{figure}

We use \llamabig as the judge LLM for 8B-parameter models,
and \llamahuge as the judge for 70B-parameter models.
Our judge LLMs are one size larger than the reference model to get a more accurate judgement than the reference model could provide while minimizing inference costs.
We find that inter-method trends generally hold across model sizes, but warn that raw scores may not be directly comparable across sizes since different judges may be more or less strict when deciding, for example, whether a generated answer sufficiently matches the gold answer.

Since our method trains the model to refuse to answer when it is likely to get the answer wrong, we are interested in the model's ability to answer attempted questions correctly (precision) and its ability to attempt questions it can answer correctly (recall).
See Figure~\ref{fig:diagnostic_testing} for a visualization of these categories.
We measure precision with \emph{answered accuracy}, i.e., \(\text{\# correct}/\text{\# attempted}\).
Measuring recall (\(\text{\# correct}/\text{\# answerable}\)) is trickier 
because we do not have direct counterfactual information about which questions the model \emph{would} get correct if it were to attempt to answer (we call this the \textit{counterfactual accuracy}). 
As a proxy, we can count the number of \textit{false refusals} (the number of refused questions that had relevant retrievals), then measure recall as 
\(\text{\# correct}/(\text{\# correct} + \text{\# false refusals})\).

\section{Analysis and Results}

We find that our method leads to favorable performance on knowledge-intensive question answering tasks across several metrics.
Our method has higher precision (accuracy on answered questions),
higher recall (successful attempts on answerable questions), and lower counterfactual accuracy, 
compared to other methods.
In ablations, we also find that our method leads to minimal degradation in non-RAG QA settings compared to all other methods,
and achieves the highest performance (precision) across different numbers of retrievals.

Refer to \cref{sec:more-results} for per-evaluation set breakdowns of the aggregated metrics (e.g., precision, recall) in this section. 

\begin{table}
  \centering
  \small
  \input{tab/prec_rec_f1}
  \caption{Summary results for precision (accuracy on attempted questions), recall (accuracy on answerable questions) and F1, the harmonic mean of both. \ours methods (ours) consistently outperform baselines.} 
  \label{tab:prec_rec_f1}
\end{table}

\subsection{Precision}
\label{sec:precision}

\begin{table*}
\centering
\footnotesize
\renewcommand{\arraystretch}{1.25}
  \input{tab/true_refusals}
  \caption{For many inputs, \baseline attempts and hallucinates an answer while \ours refuses.}
  \label{tab:refusal_examples}
\end{table*}

\begin{table*}
  \centering
  \small
  \input{tab/accuracy}
  \caption{Accuracy results are mixed across datasets and training strategies. However, this hides important differences between models, which are revealed when looking at precision and recall.
  }
  \label{tab:accuracy}
\end{table*}

As shown in the first column of \cref{tab:prec_rec_f1}, \textbf{training on self-demos leads to accuracy gains for answered questions} 
(see \cref{tab:correctness_rate} for per-dataset results.)
In other words, models trained with \ours are more likely than other baselines to answer correctly when attempting to answer. 	
We attribute this improved answered accuracy to our model's superior ability to identify and refuse questions it is likely to get wrong. For concrete examples of this, see \cref{tab:refusal_examples}. 
The other possible explanation would be that \ours causes the model to increase the total number of correct answers without increasing the number of incorrect ones, but \cref{tab:accuracy} rules this out by showing that the total number of correct answers does not systematically increase for any training strategy. 

To further support the hypothesis that \ours teaches the model to refuse questions it will likely get wrong, 
we would like to know the proportion of refused questions that the model \emph{could} have answered correctly, the lower the better.
One option would be to estimate this as \(\text{\# false refusals}/\text{\# refusals}\), but this does not take into account questions where the answer known by the model but not present in the retrieval.
To guard against this, we also estimate the model's counterfactual accuracy by checking the accuracy of the reference model on the refused instances.
These two metrics turn out to be highly correlated,
as shown in \cref{tab:counterfactual_metrics}, 
and we find that the counterfactual accuracy on refused instances for \ours is consistently lowest out of all models across model sizes,
indicating that the \textbf{answered accuracy gains are due to the \ours model refusing questions that it will likely get wrong.}

\begin{table*}
  \centering
  \small
  \input{tab/refuse}
  \caption{
    Change in refusal rates compared to the reference model (absolute change in percent) for different methods across evaluation sets. 
    Most model refusal rates do not change much overall, with the exception that
    instruction tuning 8B without retreivals (IT) decreases refusals,
    while \ours on the 70B model increases refusals.
  }
  \label{tab:refusals}
\end{table*}

Though \ours achieves the highest precision among the baselines, 
looking at differences in other metrics suggests that \textbf{different \ours models achieve high precision in different ways.}
In particular, 70B \ours (without DPO)
gets slightly fewer total answers correct compared to the \baseline model (see \cref{tab:accuracy}).
This is driven by an increase in refusals, shown in Table~\ref{tab:refusals}, which include questions the model would likely answer correctly if attempted.
The model therefore achieves high precision by attempting only high-confidence questions.
On the other hand, \ours models that do not show significant refusal increase 
must achieve higher precision through a more balanced combination of refusing questions it will get wrong,
using the context to correctly answer more questions,
and answering additional questions it would have otherwise refused. 

\begin{table*}
\centering
\footnotesize
  \begin{threeparttable}
  \renewcommand{\arraystretch}{1.25}
  \input{tab/false_refusals}
    \begin{tablenotes}
    \item[a] Some names and titles have been replaced with fictional ones.
    \end{tablenotes}
  \caption{Example outputs from 8B models trained with \baseline and \ours. For many of the inputs, \baseline completely ignores the question instead of attempting to answer it. Though we classify these as refusals, they are of much lower quality than the direct refusals from \ours shown in Table~\ref{tab:refusal_examples}. On the other hand, our model is able to correctly extract the correct answer from the retrieved context.}
  \label{tab:answered}
  \end{threeparttable}
\end{table*}

Observing the outputs of both the \ours and the \baseline models in Table~\ref{tab:answered}, 
we find that many of the \baseline ``refusals'' are simply cases of the model completely ignoring the question.
We hypothesize that these types of answers that ignore the question may stem from summarization tasks found in the instruction tuning dataset.
If we counted these as incorrect rather than refused we would see an even bigger difference between the models' answered accuracy.
In other words, our precision gains are likely \emph{underestimates}.
The fact that our models do not suffer from this type of degenerate behavior (despite training on the same data) indicates that \textbf{\ours reduces the impact of low-quality training data.}

\subsection{Recall}
\label{sec:recall}

\cref{tab:prec_rec_f1} shows that \ours outperforms all other models on recall, 
i.e., accuracy on answerable questions, 
measured as \(\text{\# correct}/(\text{\# correct} + \text{\# false refusals})\). 
We attribute this to the fact that \ours reduces false refusals compared to \baseline, as discussed in the previous section
(\cref{tab:false_refusals_with_baseline} gives a breakdown of false refusals by dataset.) 
This supports our hypothesis that training on self-demos teaches the model to better incorporate relevant context compared to \baseline.

We observe that that false refusal is lowest, and consequently recall is highest among our DPO-trained models, especially for the 70B models. 
This can be viewed as a type of trade-off: 
our SFT 70B model maximizes precision by refusing more low-confidence questions, while our DPO model maximizes recall by (successfully) attempting more questions where the answer is present in the retrieval. 
Both strategies result in high F1 scores.

\begin{table}
  \centering
  \small
  \input{tab/counterfactual_metrics.tex}
  \caption{Reference model accuracy on refused questions and false refusal rates. Lower is better. These two measures of counterfactual accuracy are highly correlated.}
  \label{tab:counterfactual_metrics}
\end{table}

\subsection{Number of retrievals}

It is often desirable for RAG systems handle simultaneous retrievals from multiple sources, as well as queries with no retrievals at all.
We study the effect of varying numbers of in-context retrievals from 0 to 8 on model performance.
For each number of retrievals~\(n\), we include the \(n\) most-relevant retrievals for the question, as scored by the retriever system.

\cref{tab:multisource} shows that all models exhibit monotonic improvement as the number of retrievals increases, 
even when the number of retrievals surpasses the number of retrievals trained on.
Across the board, \ours (with or without DPO) achieves the highest performance.

The ``0'' column of~\cref{tab:multisource},
shows that both \textbf{\baseline and IT seriously degrade the LLM's performance (measured with precision)
on QA without retrievals.}
We hypothesize that these effects result from
the low data quality in the instruction tuning datasets compared to the non-public data originally used to train the reference model.
While \ours uses the same instruction tuning dataset, the self-generated responses prevent the model from actually learning to predict the lower-quality gold responses,
better preserving the model's original distribution and causing no significant degradation in non-RAG settings. 

\begin{table}
  \centering
  \small
  \input{tab/retrievals}
  \caption{Model performance (precision) under different numbers of retrievals. When retrievals are present, \baseline models outperform non-\baseline models.}
  \label{tab:multisource}
\end{table}

\section{Discussion and conclusion}

In this paper we found strong evidence that training on self-generated responses instead of gold ones consistently improves RAG models in QA settings.
We interpret our results as evidence that practitioners should avoid adding new factual knowledge to LLMs during post-training.
Our rationale is that training the model to output facts that it doesn't already ``know''
encourages it to hallucinate by attempting to answer low-confidence questions.
Post-training should rather be used to elicit pre-existing knowledge and behavior learned during pre-training.

Our second major conclusion is that \ours enables successful training on low-quality datasets.
Artifacts such as summarization tasks in the training data mean that na\"ive instruction tuning methods degrade model behavior. 
By training on self-demos, \ours avoids teaching models to generate low-quality outputs while still allowing the model to benefit from the supervision and task adaptation aspects of the training data. 

Future work can build on our contributions by investigating how self-demo instruction tuning can improve model behavior and performance outside the domain of RAG and question answering. 
We are also interested in techniques that control the trade-off between precision and recall that we saw between 70B SFT and DPO models in \cref{sec:recall}.

\section{Limitations}

Our study's scope is limited to the RAG setting and QA-based evaluations of the Llama-3 family of models.
Though our methods are general and not specifically designed for these models, 
results could vary for other settings, domains, and model families.

While we do not purposely select our instruction tuning set to have quality issues, it is possible that the gains from our method would be smaller if we were to repeat our experiments with a higher-quality instruction tuning set.

\bibliography{main}

\appendix

\section{Fine-grained results}
\label{sec:more-results}

\cref{tab:answerable_accuracy,tab:counterfactual,tab:false_refusals_with_baseline,tab:f1,tab:correctness_rate}
break down the aggregated metrics from the main section of the paper into per-evaluation-set results. 

\bigskip
\noindent Precision \dotfill \cref{tab:correctness_rate} \\
Recall \dotfill \cref{tab:answerable_accuracy} \\
F1 \dotfill \cref{tab:f1} \\
False refusals \dotfill \cref{tab:false_refusals_with_baseline} \\
Counteractual accuracy \dotfill \cref{tab:counterfactual} \\

\begin{table*}
  \centering
  \small
  \input{tab/correctness_rate}
  \caption{Precision, as measured by accuracy on answered (not refused) questions, for all models and evaluation sets. \llamasmall fine-tuned with \ours consistently achieves higher precision (i.e., \(\text{\# correct}/\text{\# answered}\)) compared to all other methods, with \ours + DPO performing the best on average.}
  \label{tab:correctness_rate}
\end{table*}

\begin{table*}
  \centering
  \small
  \input{tab/answerable_accuracy}
  \caption{
    Recall, as measured by proportion of questions answered correctly when the answer is contained in the retrieved context (i.e., \textit{answerable accuracy}) for each model and evaluation set.
    For 8B-parameter models, recall is highest for the \ours methods.
    For 70B-parameter models, \ours sees less recall degradation compared to \baseline.
  }
  \label{tab:answerable_accuracy}
\end{table*}

\begin{table*}
  \centering
  \small
  \input{tab/f1}
  \caption{
    F1-score, the harmonic mean of precision and recall, for each model and dataset.
    F1 is a measure of performance that takes into account both the model's ability to answer attempted questions correctly, and its ability to attempt and correctly answer questions when the answer is present in the retrieved context.
    On average and across sizes, \ours methods achieves the highest F1 score, demonstrating the superiority of training on self-generated demonstrations over gold labels.
  }
  \label{tab:f1}
\end{table*}

\begin{table*}
  \centering
  \small
  \input{tab/false_refusals_with_baseline}
  \caption{False refusal rates for \baseline and \ours. Lower is better. For the 8B model size, \ours consistently exhibits fewer false refusals than \baseline.}
  \label{tab:false_refusals_with_baseline}
\end{table*}

\begin{table*}
  \centering
  \small
  \input{tab/counterfactual}
  \caption{
    Counterfactual accuracy on refused questions (as estimated by the reference model, lower is better) for \llamasmall. Across datasets \ours (both SFT and DPO) consistently have lowest counterfactual accuracy, indicating that they are better at choosing questions to refuse.
	}
  \label{tab:counterfactual}
\end{table*}

\end{document}

%% file: abstract.tex
Large language models (LLMs) often struggle with knowledge intensive NLP tasks,
such as answering ``Who won the latest World Cup?''
because the knowledge they learn during training may be insufficient or outdated.
Conditioning generation on retrieved documents---a technique known as retrieval augmented generation (RAG)---mitigates these shortcomings by allowing the model to leverage in-context information.
Practitioners can improve LLM RAG performance 
by fine-tuning on retrieval-augmented instructions, 
but must beware that this can cause undesirable model behaviors like hallucinations. 
We attribute this degradation to the fact that the training data is likely to be out-of-distribution for the model and may suffer from quality issues, 
such as misalignment between retrievals and target responses  
(since retrievals are frequently added post-hoc).
We propose a recipe for training RAG-enabled LLMs
using self-generated demonstrations,
thereby avoiding training on out-of-distribution text and integrating retrievals into the LLM responses.
We evaluate our method on knowledge intensive question answering (QA) tasks and show that
our method teaches LLMs to properly handle in-context retrievals and abstain from questions it will likely get wrong.
Compared to conventional RA-IT methods, our method prevents model degradation in non-RAG settings while exhibiting superior QA performance.

%% file: fig/baseline.tex
\begin{tikzpicture}[node distance=8mm, rounded corners]
  \node[draw] (pretrain) {Instruct model};
  \node[draw, above=of pretrain] (instruct) {Retrieval-tuned instruct model}
    edge[latex-] node[right, font=\smaller, near start] (sft) {Train on \sethlcolor{pink}\hl{OOD} data} (pretrain);
  \node[draw, below=of pretrain] (ours) {Self-demo+retrieval-tuned instruct model (ours)}
    edge[latex-] node[right, font=\smaller, near start] (sdrait) {Train on \sethlcolor{green}\hl{in-distribution} data} (pretrain);
  \node[left=2.5 of instruct] (rair) {Retrieval-augmented instruction-response pairs};
  \node[
    below=5mm of rair,
    text width=2cm,
    label=above:Instruction,
    draw,
    font=\smaller,
  ] (question) {
    What color is the roof of the Frederick Krause Mini Mansion? 
  };
  \node[
    left=1mm of question.north west,
    text width=2cm,
    label=above:Retrieval,
    draw,
    anchor=north east,
    font=\smaller,
  ] (retrieval) {
    The Frederick Krause Mini Mansion is a historic\ldots
  };
  \node[
    right=1mm of question.north east, 
    anchor=north west,
    text width=2cm,
    fill=pink,
    label=above:Response,
    draw,
    font=\smaller,
  ] (answer) {
    Grey and pink
  };
  \node[below=1mm of answer, fill=green, text width=2cm, draw, font=\smaller] (selfdemo) {According to the passage, the colors are gray-blue and rose.};
  \node[draw, fit={(rair) (question) (retrieval) (answer) (selfdemo)}] (data) {};
  \draw (answer.east) to[out=north east, in=south, looseness=0.5] (sft.west);
  \draw[-latex] (pretrain) to[out=west, in=north east] node[below, font=\smaller, fill=white] {Generate \& filter} (selfdemo.east);
  \draw (selfdemo.east) to[out=south east, in=north, looseness=0.5] (sdrait.west);
\end{tikzpicture}

%% file: fig/method.tex
\begin{tikzpicture}[node distance=8mm, rounded corners, text height=1.5ex, text depth=.25ex]
  \node[draw] (pretrain) {Instruct model};
  \node[draw] (candidates) [right=18mm of pretrain] {Response candidates} 
    edge[latex-] node[below, font=\smaller] (gen) {Generate} (pretrain);
  \node[draw] (instruct) [below=15mm of pretrain, text width=2cm, text height=, text depth=] {Self-demo retrieval-tuned instruct model}
    edge[latex-] node[left, near start, text width=2cm, font=\smaller, text height=, text depth=] (sft) {Train on \sethlcolor{green}\hl{in-distribution} data} (pretrain);
  \node[right=of candidates] (label) {Retrieval augmented instruction-response pairs};
  \node[below=1mm of label, draw] (instructions) {Instructions};
  \node[left=1mm of instructions, draw] (retrievals) {Retrievals};
  \node[right=1mm of instructions, draw, fill=pink] (responses) {Responses};
  \node[draw, fit=(retrievals) (instructions) (responses) (label), text width=7cm] (training set) {}; 
  \draw (training set) to[out=160, in=west, looseness=1] node[fill=white, rounded corners, font=\smaller] {Retrievals and instructions only} (gen.north);
  \node[draw, below=15mm of candidates, fill=green] (demos) {Self-generated demos} 
    edge[latex-] node[near start, left, font=\smaller] (filter) {Filter} (candidates);
  \draw (pretrain) to[out=-30, in=north, looseness=0.5] node[below, font=\smaller] {LLM-as-judge} (filter.north east);
  \draw (training set) to[out=west, in=north, looseness=0.5] (filter.north east);
  \coordinate[below=5mm of demos, label={[font=\smaller]below:Replace responses with self-demos}] (combine);
  \draw (demos) to[out=south, in=east] (combine);
  \draw (training set) to[out=200, in=east] (combine) to[out=west, in=north] (sft.north east);
\end{tikzpicture}

%% file: tab/prec_rec_f1.tex
\begin{tabular}{llSSS}
\toprule
{} & {} & {Precision} & {Recall} & {F1} \\
\midrule
\multirow[c]{5}{*}{8B} & \llamainstruct & \cellcolor[rgb]{0.9390142252979623, 0.9755601691657055, 0.9339730872741253} 75.9 & \cellcolor[rgb]{0.9937254901960785, 0.9976470588235294, 0.9921568627450981} 78.4 & \cellcolor[rgb]{0.9826528258362168, 0.9933410226835833, 0.9792387543252595} 76.9 \\
 & IT & \cellcolor[rgb]{0.9937254901960785, 0.9976470588235294, 0.9921568627450981} 73.4 & \cellcolor[rgb]{0.9258592848904268, 0.9700023068050749, 0.9212272202998846} 79.9 & \cellcolor[rgb]{0.9937254901960785, 0.9976470588235294, 0.9921568627450981} 76.4 \\
 & RA-IT & \cellcolor[rgb]{0.826805074971165, 0.9084659746251441, 0.8536562860438293} 79.2 & \cellcolor[rgb]{0.9103575547866205, 0.9627681660899654, 0.908481353325644} 80.2 & \cellcolor[rgb]{0.8514509803921568, 0.9343483275663207, 0.8731718569780853} 79.6 \\
 & \ours & \cellcolor[rgb]{0.8, 0.8674571318723567, 0.8270326797385621} 80.6 & \cellcolor[rgb]{0.8, 0.8563598615916955, 0.8224313725490195} 82.3 & \cellcolor[rgb]{0.8, 0.8644306036139946, 0.8257777777777777} 81.3 \\
 & \ours DPO & \cellcolor[rgb]{0.8, 0.8533333333333333, 0.8211764705882353} 81.1 & \cellcolor[rgb]{0.8, 0.8533333333333333, 0.8211764705882353} 82.4 & \cellcolor[rgb]{0.8, 0.8533333333333333, 0.8211764705882353} 81.5 \\
\hhline{*{5}{=}}
\multirow[c]{5}{*}{70B} & \llamainstruct & \cellcolor[rgb]{0.9799953863898501, 0.9923075740099961, 0.9761384083044983} 73.8 & \cellcolor[rgb]{0.9492995001922337, 0.9798908112264514, 0.9439876970396002} 79.4 & \cellcolor[rgb]{0.9539746251441753, 0.9818592848904267, 0.9485397923875433} 76.4 \\
 & IT & \cellcolor[rgb]{0.9937254901960785, 0.9976470588235294, 0.9921568627450981} 72.9 & \cellcolor[rgb]{0.9937254901960785, 0.9976470588235294, 0.9921568627450981} 77.3 & \cellcolor[rgb]{0.9937254901960785, 0.9976470588235294, 0.9921568627450981} 74.9 \\
 & RA-IT & \cellcolor[rgb]{0.9399492502883506, 0.9759538638985006, 0.934883506343714} 75.2 & \cellcolor[rgb]{0.9901822376009227, 0.9962691272587466, 0.9880230680507497} 77.5 & \cellcolor[rgb]{0.9581084198385237, 0.9835570934256055, 0.9527535563244905} 76.3 \\
 & \ours & \cellcolor[rgb]{0.8, 0.8533333333333333, 0.8211764705882353} 80.1 & \cellcolor[rgb]{0.955844675124952, 0.9826466743560169, 0.9503606305267205} 79.2 & \cellcolor[rgb]{0.8207766243752403, 0.9032987312572087, 0.8493502499038831} 79.4 \\
 & \ours DPO & \cellcolor[rgb]{0.8614901960784314, 0.9392695117262592, 0.8780930411380239} 77.1 & \cellcolor[rgb]{0.8, 0.8533333333333333, 0.8211764705882353} 84.8 & \cellcolor[rgb]{0.8, 0.8533333333333333, 0.8211764705882353} 80.7 \\
\cline{1-5}
\bottomrule
\end{tabular}

%% file: tab/true_refusals.tex
\sethlcolor{pink}
\begin{tabularx}\linewidth{p{0.22\linewidth}XX}
  \toprule
  Question (Answer) & \baseline & \ours (ours) \\
  \midrule
  Who sings ``Echo Flux''?
  (Aurora Wychwood)
  & The song ``Echo Flux'' is performed by American country music artist \hl{Kaius Riven}. 
  & To answer your question, I'd need more context \ldots{} I recommend checking online lyrics or a music database for more information. \\
When did england last lose three in a row?
(27/05/2018)
  & The last time England lost three consecutive matches in international football was \hl{between 1999 and 2000}. 
  & To answer your question, I'd need to review England's match history. \\
Who is Hawk in Beyond the Red Horizon?
  (Asher Wilder)
  & Hawk is a fictional character in the film ``Beyond the Red Horizon'' \ldots{} played by actor \hl{Ryder Blackwood}. 
  & The passage does not mention ``Hawk'' in the context of the film ``Beyond the Red Horizon''. \\
\bottomrule
\end{tabularx}

%

%% file: tab/accuracy.tex
\begin{tabular}{llS|SSSSSSSSS}
\toprule
{} & {} & {\textbf{Avg.}} & {PSR} & {FEVER} & {HPQA} & {MMLU} & {NQ} & {TQA} & {T-REx} & {WoW} & {zsRE} \\
\midrule
\multirow[c]{5}{*}{8B} & \llamainstruct & \cellcolor[rgb]{0.9937254901960785, 0.9976470588235294, 0.9921568627450981} 66.1 & \cellcolor[rgb]{0.9937254901960785, 0.9976470588235294, 0.9921568627450981} 60.3 & \cellcolor[rgb]{0.880313725490196, 0.9484967320261438, 0.8873202614379085} 78.9 & \cellcolor[rgb]{0.9879677047289505, 0.9954079200307574, 0.985439446366782} 52.6 & \cellcolor[rgb]{0.8, 0.8533333333333333, 0.8211764705882353} 79.5 & \cellcolor[rgb]{0.8291118800461361, 0.9107912341407151, 0.8554463667820069} 64.4 & \cellcolor[rgb]{0.9937254901960785, 0.9976470588235294, 0.9921568627450981} 74.7 & \cellcolor[rgb]{0.9937254901960785, 0.9976470588235294, 0.9921568627450981} 64.2 & \cellcolor[rgb]{0.9937254901960785, 0.9976470588235294, 0.9921568627450981} 45.0 & \cellcolor[rgb]{0.9826528258362168, 0.9933410226835833, 0.9792387543252595} 75.2 \\
 & IT & \cellcolor[rgb]{0.8752941176470588, 0.9460361399461745, 0.8848596693579392} 68.8 & \cellcolor[rgb]{0.8514509803921568, 0.9343483275663207, 0.8731718569780853} 70.7 & \cellcolor[rgb]{0.9937254901960785, 0.9976470588235294, 0.9921568627450981} 68.7 & \cellcolor[rgb]{0.8069973087274125, 0.8914878892733564, 0.8395078815840061} 56.5 & \cellcolor[rgb]{0.9937254901960785, 0.9976470588235294, 0.9921568627450981} 73.1 & \cellcolor[rgb]{0.9937254901960785, 0.9976470588235294, 0.9921568627450981} 57.0 & \cellcolor[rgb]{0.8298500576701269, 0.9115786236063053, 0.8560369088811995} 77.7 & \cellcolor[rgb]{0.8, 0.8533333333333333, 0.8211764705882353} 72.4 & \cellcolor[rgb]{0.8216378316032296, 0.9040369088811995, 0.8499653979238755} 62.8 & \cellcolor[rgb]{0.8, 0.8533333333333333, 0.8211764705882353} 80.2 \\
 & RA-IT & \cellcolor[rgb]{0.8, 0.8674571318723567, 0.8270326797385621} 70.7 & \cellcolor[rgb]{0.8, 0.8533333333333333, 0.8211764705882353} 77.0 & \cellcolor[rgb]{0.8948558246828143, 0.9555340253748559, 0.8957354863514033} 78.0 & \cellcolor[rgb]{0.9937254901960785, 0.9976470588235294, 0.9921568627450981} 52.3 & \cellcolor[rgb]{0.9595847750865052, 0.9841476355247981, 0.9543283352556708} 74.6 & \cellcolor[rgb]{0.8970703575547866, 0.9565674740484429, 0.8975563244905805} 61.8 & \cellcolor[rgb]{0.8, 0.8533333333333333, 0.8211764705882353} 78.7 & \cellcolor[rgb]{0.8665098039215686, 0.9417301038062283, 0.8805536332179931} 68.9 & \cellcolor[rgb]{0.8, 0.8533333333333333, 0.8211764705882353} 68.0 & \cellcolor[rgb]{0.9014994232987312, 0.958634371395617, 0.9011980007689351} 77.2 \\
 & \ours & \cellcolor[rgb]{0.8, 0.8533333333333333, 0.8211764705882353} 71.0 & \cellcolor[rgb]{0.8001076509034987, 0.8855824682814302, 0.8345866974240677} 74.9 & \cellcolor[rgb]{0.8320645905420991, 0.9139407920030758, 0.8578085351787774} 82.5 & \cellcolor[rgb]{0.8, 0.8533333333333333, 0.8211764705882353} 57.2 & \cellcolor[rgb]{0.8401845444059977, 0.9226020761245675, 0.8643044982698962} 77.5 & \cellcolor[rgb]{0.8, 0.8583775470972703, 0.8232679738562092} 66.7 & \cellcolor[rgb]{0.8614901960784314, 0.9392695117262592, 0.8780930411380239} 77.1 & \cellcolor[rgb]{0.9059284890426759, 0.9607012687427913, 0.9048396770472895} 67.8 & \cellcolor[rgb]{0.8627450980392157, 0.9398846597462515, 0.8787081891580162} 58.5 & \cellcolor[rgb]{0.9287289504036909, 0.9712295271049596, 0.9239584775086505} 76.6 \\
 & \ours DPO & \cellcolor[rgb]{0.8259438677431756, 0.9077277970011534, 0.853041138023837} 69.8 & \cellcolor[rgb]{0.9158938869665513, 0.9653517877739332, 0.913033448673587} 67.2 & \cellcolor[rgb]{0.8, 0.8533333333333333, 0.8211764705882353} 87.8 & \cellcolor[rgb]{0.8564705882352941, 0.9368089196462899, 0.8756324490580546} 55.3 & \cellcolor[rgb]{0.8078585159554017, 0.8922260668973472, 0.8401230296039984} 78.5 & \cellcolor[rgb]{0.8, 0.8533333333333333, 0.8211764705882353} 67.0 & \cellcolor[rgb]{0.9114648212226066, 0.9632848904267589, 0.9093917723952326} 76.4 & \cellcolor[rgb]{0.9923967704728951, 0.9971303344867358, 0.9906066897347174} 64.3 & \cellcolor[rgb]{0.8815686274509804, 0.9491118800461361, 0.8879354094579008} 57.1 & \cellcolor[rgb]{0.9937254901960785, 0.9976470588235294, 0.9921568627450981} 74.6 \\
\hhline{*{12}{=}}
\multirow[c]{5}{*}{70B} & \llamainstruct & \cellcolor[rgb]{0.9844244521337947, 0.9940299884659747, 0.9813056516724337} 68.2 & \cellcolor[rgb]{0.9758246828143022, 0.990643598615917, 0.9716509034986544} 61.2 & \cellcolor[rgb]{0.8, 0.8624129181084198, 0.8249411764705883} 90.4 & \cellcolor[rgb]{0.9654901960784313, 0.9865098039215686, 0.9606274509803922} 56.6 & \cellcolor[rgb]{0.8, 0.8533333333333333, 0.8211764705882353} 76.8 & \cellcolor[rgb]{0.8147481737793156, 0.8981314878892733, 0.8450442137639369} 67.6 & \cellcolor[rgb]{0.9937254901960785, 0.9976470588235294, 0.9921568627450981} 79.5 & \cellcolor[rgb]{0.8981776239907728, 0.9570841983852365, 0.8984667435601692} 71.3 & \cellcolor[rgb]{0.9937254901960785, 0.9976470588235294, 0.9921568627450981} 29.1 & \cellcolor[rgb]{0.8514509803921568, 0.9343483275663207, 0.8731718569780853} 81.3 \\
 & IT & \cellcolor[rgb]{0.9937254901960785, 0.9976470588235294, 0.9921568627450981} 67.8 & \cellcolor[rgb]{0.9937254901960785, 0.9976470588235294, 0.9921568627450981} 59.5 & \cellcolor[rgb]{0.9937254901960785, 0.9976470588235294, 0.9921568627450981} 70.5 & \cellcolor[rgb]{0.8156093810073048, 0.8988696655132641, 0.8456593617839292} 61.6 & \cellcolor[rgb]{0.9371441753171857, 0.9747727797001153, 0.9321522491349481} 72.2 & \cellcolor[rgb]{0.8715294117647059, 0.9441906958861976, 0.8830142252979623} 64.1 & \cellcolor[rgb]{0.8815686274509804, 0.9491118800461361, 0.8879354094579008} 83.4 & \cellcolor[rgb]{0.8552156862745098, 0.9361937716262976, 0.8750173010380623} 72.9 & \cellcolor[rgb]{0.8350173010380623, 0.9170903498654364, 0.8601707035755478} 44.5 & \cellcolor[rgb]{0.8490426758938869, 0.9320507497116494, 0.8713910034602076} 81.4 \\
 & RA-IT & \cellcolor[rgb]{0.9059284890426759, 0.9607012687427913, 0.9048396770472895} 70.0 & \cellcolor[rgb]{0.8970703575547866, 0.9565674740484429, 0.8975563244905805} 65.1 & \cellcolor[rgb]{0.8828235294117647, 0.9497270280661284, 0.8885505574778931} 81.3 & \cellcolor[rgb]{0.8, 0.8533333333333333, 0.8211764705882353} 63.2 & \cellcolor[rgb]{0.8250826605151864, 0.9069896193771626, 0.8524259900038447} 75.2 & \cellcolor[rgb]{0.9937254901960785, 0.9976470588235294, 0.9921568627450981} 56.3 & \cellcolor[rgb]{0.8, 0.8533333333333333, 0.8211764705882353} 86.9 & \cellcolor[rgb]{0.8853333333333333, 0.9509573241061131, 0.8897808535178777} 71.8 & \cellcolor[rgb]{0.8, 0.8543421760861207, 0.8215947712418301} 50.7 & \cellcolor[rgb]{0.9092502883506344, 0.9622514417531719, 0.9075709342560554} 79.5 \\
 & \ours & \cellcolor[rgb]{0.9910680507497116, 0.9966136101499423, 0.9890565167243368} 67.9 & \cellcolor[rgb]{0.8, 0.8533333333333333, 0.8211764705882353} 71.3 & \cellcolor[rgb]{0.9511695501730104, 0.9806782006920415, 0.9458085351787774} 76.2 & \cellcolor[rgb]{0.9315340253748559, 0.9724106113033448, 0.9266897347174163} 57.9 & \cellcolor[rgb]{0.9937254901960785, 0.9976470588235294, 0.9921568627450981} 69.9 & \cellcolor[rgb]{0.823360246059208, 0.9055132641291811, 0.8511956939638601} 67.0 & \cellcolor[rgb]{0.9923967704728951, 0.9971303344867358, 0.9906066897347174} 79.6 & \cellcolor[rgb]{0.9937254901960785, 0.9976470588235294, 0.9921568627450981} 65.7 & \cellcolor[rgb]{0.8, 0.8805720876585929, 0.8324705882352941} 48.5 & \cellcolor[rgb]{0.9937254901960785, 0.9976470588235294, 0.9921568627450981} 75.2 \\
 & \ours DPO & \cellcolor[rgb]{0.8, 0.8533333333333333, 0.8211764705882353} 72.7 & \cellcolor[rgb]{0.8250826605151864, 0.9069896193771626, 0.8524259900038447} 68.4 & \cellcolor[rgb]{0.8, 0.8533333333333333, 0.8211764705882353} 91.1 & \cellcolor[rgb]{0.9937254901960785, 0.9976470588235294, 0.9921568627450981} 54.9 & \cellcolor[rgb]{0.8, 0.860395232602845, 0.8241045751633986} 76.6 & \cellcolor[rgb]{0.8, 0.8533333333333333, 0.8211764705882353} 70.3 & \cellcolor[rgb]{0.9897393310265282, 0.9960968858131488, 0.9875063437139562} 79.8 & \cellcolor[rgb]{0.8, 0.8533333333333333, 0.8211764705882353} 77.5 & \cellcolor[rgb]{0.8, 0.8533333333333333, 0.8211764705882353} 50.8 & \cellcolor[rgb]{0.8, 0.8533333333333333, 0.8211764705882353} 85.0 \\
\cline{1-12}
\bottomrule
\end{tabular}

%% file: tab/refuse.tex
\begin{tabular}{llS[retain-explicit-plus]|S[table-format=2, retain-explicit-plus]S[table-format=2, retain-explicit-plus]S[table-format=2, retain-explicit-plus]S[table-format=2, retain-explicit-plus]S[table-format=2, retain-explicit-plus]S[table-format=2, retain-explicit-plus]S[table-format=2, retain-explicit-plus]S[table-format=2, retain-explicit-plus]S[table-format=2, retain-explicit-plus]}
\toprule
{} & {} & {\textbf{Avg.}} & {PSR} & {FEVER} & {HPQA} & {MMLU} & {NQ} & {TQA} & {T-REx} & {WoW} & {zsRE} \\
\midrule
\multirow[c]{4}{*}{8B} & IT & \cellcolor[rgb]{0.8603921568627451, 0.8603921568627451, 1.0} -7 & \cellcolor[rgb]{0.9043137254901961, 0.9043137254901961, 1.0} -5 & \cellcolor[rgb]{0.9984313725490196, 0.9984313725490196, 1.0} -0 & \cellcolor[rgb]{0.8, 0.8, 0.9588235294117646} -13 & \cellcolor[rgb]{1.0, 0.9545098039215686, 0.9545098039215686} +2 & \cellcolor[rgb]{0.9388235294117647, 0.9388235294117647, 1.0} -3 & \cellcolor[rgb]{0.9294117647058824, 0.9294117647058824, 1.0} -4 & \cellcolor[rgb]{0.8, 0.8, 0.9434509803921569} -14 & \cellcolor[rgb]{0.8, 0.8, 0.9017254901960784} -17 & \cellcolor[rgb]{0.8, 0.8, 0.9983529411764706} -10 \\
 & RA-IT & \cellcolor[rgb]{0.9513725490196079, 0.9513725490196079, 1.0} -2 & \cellcolor[rgb]{0.8949019607843137, 0.8949019607843137, 1.0} -5 & \cellcolor[rgb]{1.0, 0.9670588235294117, 0.9670588235294117} +2 & \cellcolor[rgb]{1.0, 0.8949019607843137, 0.8949019607843137} +5 & \cellcolor[rgb]{1.0, 0.8980392156862745, 0.8980392156862745} +5 & \cellcolor[rgb]{0.9639215686274509, 0.9639215686274509, 1.0} -2 & \cellcolor[rgb]{1.0, 0.9921568627450981, 0.9921568627450981} +0 & \cellcolor[rgb]{0.8949019607843137, 0.8949019607843137, 1.0} -5 & \cellcolor[rgb]{0.8, 0.8, 0.8643921568627451} -20 & \cellcolor[rgb]{0.9419607843137255, 0.9419607843137255, 1.0} -3 \\
 & \ours & \cellcolor[rgb]{0.9733333333333334, 0.9733333333333334, 1.0} -1 & \cellcolor[rgb]{0.9043137254901961, 0.9043137254901961, 1.0} -5 & \cellcolor[rgb]{0.9545098039215686, 0.9545098039215686, 1.0} -2 & \cellcolor[rgb]{0.9294117647058824, 0.9294117647058824, 1.0} -3 & \cellcolor[rgb]{1.0, 0.9482352941176471, 0.9482352941176471} +3 & \cellcolor[rgb]{1.0, 0.9670588235294117, 0.9670588235294117} +2 & \cellcolor[rgb]{1.0, 0.9764705882352941, 0.9764705882352941} +1 & \cellcolor[rgb]{0.9670588235294117, 0.9670588235294117, 1.0} -2 & \cellcolor[rgb]{0.9388235294117647, 0.9388235294117647, 1.0} -3 & \cellcolor[rgb]{0.9670588235294117, 0.9670588235294117, 1.0} -2 \\
 & \ours DPO & \cellcolor[rgb]{1.0, 0.9858823529411765, 0.9858823529411765} +1 & \cellcolor[rgb]{0.9545098039215686, 0.9545098039215686, 1.0} -2 & \cellcolor[rgb]{0.9074509803921568, 0.9074509803921568, 1.0} -5 & \cellcolor[rgb]{1.0, 0.9294117647058824, 0.9294117647058824} +4 & \cellcolor[rgb]{1.0, 0.8729411764705882, 0.8729411764705882} +6 & \cellcolor[rgb]{1.0, 0.9137254901960784, 0.9137254901960784} +4 & \cellcolor[rgb]{1.0, 0.9482352941176471, 0.9482352941176471} +3 & \cellcolor[rgb]{1.0, 0.9168627450980392, 0.9168627450980392} +4 & \cellcolor[rgb]{0.8, 0.8, 0.9961568627450981} -10 & \cellcolor[rgb]{1.0, 0.9388235294117647, 0.9388235294117647} +3 \\
\hhline{*{12}{=}}
\multirow[c]{4}{*}{70B} & IT & \cellcolor[rgb]{0.9733333333333334, 0.9733333333333334, 1.0} -1 & \cellcolor[rgb]{1.0, 0.9513725490196079, 0.9513725490196079} +2 & \cellcolor[rgb]{0.951764705882353, 0.8, 0.8} +15 & \cellcolor[rgb]{0.8227450980392157, 0.8227450980392157, 1.0} -9 & \cellcolor[rgb]{0.9921568627450981, 0.9921568627450981, 1.0} -0 & \cellcolor[rgb]{0.9545098039215686, 0.9545098039215686, 1.0} -2 & \cellcolor[rgb]{0.9639215686274509, 0.9639215686274509, 1.0} -2 & \cellcolor[rgb]{0.9105882352941177, 0.9105882352941177, 1.0} -4 & \cellcolor[rgb]{0.8164705882352941, 0.8164705882352941, 1.0} -9 & \cellcolor[rgb]{0.9513725490196079, 0.9513725490196079, 1.0} -2 \\
 & RA-IT & \cellcolor[rgb]{0.9733333333333334, 0.9733333333333334, 1.0} -1 & \cellcolor[rgb]{1.0, 0.9858823529411765, 0.9858823529411765} +1 & \cellcolor[rgb]{1.0, 0.8854901960784315, 0.8854901960784315} +6 & \cellcolor[rgb]{0.8572549019607842, 0.8572549019607842, 1.0} -7 & \cellcolor[rgb]{1.0, 0.9796078431372549, 0.9796078431372549} +1 & \cellcolor[rgb]{1.0, 0.9733333333333334, 0.9733333333333334} +1 & \cellcolor[rgb]{0.9199999999999999, 0.9199999999999999, 1.0} -4 & \cellcolor[rgb]{1.0, 0.9890196078431372, 0.9890196078431372} +1 & \cellcolor[rgb]{0.8, 0.8, 0.9785882352941176} -11 & \cellcolor[rgb]{1.0, 0.9921568627450981, 0.9921568627450981} +0 \\
 & \ours & \cellcolor[rgb]{1.0, 0.8635294117647059, 0.8635294117647059} +7 & \cellcolor[rgb]{1.0, 0.9545098039215686, 0.9545098039215686} +2 & \cellcolor[rgb]{0.9580392156862745, 0.8, 0.8} +14 & \cellcolor[rgb]{0.9921568627450981, 0.9921568627450981, 1.0} -0 & \cellcolor[rgb]{0.9925490196078431, 0.8, 0.8} +11 & \cellcolor[rgb]{1.0, 0.8572549019607842, 0.8572549019607842} +7 & \cellcolor[rgb]{1.0, 0.9419607843137254, 0.9419607843137254} +3 & \cellcolor[rgb]{0.9674509803921569, 0.8, 0.8} +13 & \cellcolor[rgb]{1.0, 0.9921568627450981, 0.9921568627450981} +0 & \cellcolor[rgb]{0.9862745098039216, 0.8, 0.8} +11 \\
 & \ours DPO & \cellcolor[rgb]{0.9545098039215686, 0.9545098039215686, 1.0} -2 & \cellcolor[rgb]{0.9890196078431372, 0.9890196078431372, 1.0} -1 & \cellcolor[rgb]{0.9921568627450981, 0.9921568627450981, 1.0} -0 & \cellcolor[rgb]{0.9074509803921568, 0.9074509803921568, 1.0} -5 & \cellcolor[rgb]{1.0, 0.9796078431372549, 0.9796078431372549} +1 & \cellcolor[rgb]{1.0, 0.9984313725490196, 0.9984313725490196} +0 & \cellcolor[rgb]{0.9701960784313726, 0.9701960784313726, 1.0} -2 & \cellcolor[rgb]{0.9325490196078431, 0.9325490196078431, 1.0} -3 & \cellcolor[rgb]{0.8196078431372549, 0.8196078431372549, 1.0} -9 & \cellcolor[rgb]{0.9545098039215686, 0.9545098039215686, 1.0} -2 \\
\cline{1-12}
\bottomrule
\end{tabular}

%% file: tab/false_refusals.tex
\begin{tabular}{p{0.241\textwidth}p{0.140\textwidth}p{0.251\textwidth}p{0.251\textwidth}}
\toprule
  Context (answer highlighted) & Question & \baseline & \ours \\
\midrule
  The accompanying music video was directed by \hl{Kaia Reyes}\tnote{a}. It premiered through \ldots 
  & Who directed ``Lost in the Haze''?  
  & ``Lost in the Haze'' is a song by \ldots\ Luna Nightingale, from her fifth studio album, ``Fractured Light''. 
  & \hl{Kaia Reyes} directed the music video for Luna Nightingale's ``Lost in the Haze''. \\
  The \$173 million Orleans opened on 88 acres of land on December 18, \hl{1996}, with \ldots 
  & When was the orleans hotel in las vegas built?
  & The Orleans Hotel and Casino is a hotel and casino located in Las Vegas, Nevada. 
  & \ldots but the Las Vegas location was opened on December 18, \hl{1996}. \\
  

  \ldots of Kerman's memoirs. The series began filming \ldots\ on \hl{March 7, 2013}. 
  & When does orange is the new black start filming?
  & 
  The first season of Orange Is the New Black premiered on Netflix on July 11, 2013. & Orange is the New Black began filming on \hl{March 7, 2013}. \\
\bottomrule
\end{tabular}

%% file: tab/counterfactual_metrics.tex
\begin{tabular}{llSS}
\toprule
{} & {} & {\shortstack{Ref.\ model acc.\ \\ on refused}} & {\shortstack{False \\ refusal rate}} \\
\midrule
\multirow[c]{4}{*}{8B} & IT & \cellcolor[rgb]{0.9937254901960785, 0.9976470588235294, 0.9921568627450981} 61.1 & \cellcolor[rgb]{0.9654901960784313, 0.9865098039215686, 0.9606274509803922} 40.9 \\
 & RA-IT & \cellcolor[rgb]{0.9736101499423299, 0.989757785467128, 0.9692887351018838} 59.3 & \cellcolor[rgb]{0.9937254901960785, 0.9976470588235294, 0.9921568627450981} 43.1 \\
 & \ours & \cellcolor[rgb]{0.8035524798154556, 0.8885351787773933, 0.8370472895040368} 51.4 & \cellcolor[rgb]{0.8350173010380623, 0.9170903498654364, 0.8601707035755478} 35.4 \\
 & \ours DPO & \cellcolor[rgb]{0.8, 0.8533333333333333, 0.8211764705882353} 49.8 & \cellcolor[rgb]{0.8, 0.8533333333333333, 0.8211764705882353} 32.3 \\
\hhline{*{4}{=}}
\multirow[c]{4}{*}{70B} & IT & \cellcolor[rgb]{0.9736101499423299, 0.989757785467128, 0.9692887351018838} 56.1 & \cellcolor[rgb]{0.9502345251826221, 0.9802845059592464, 0.9448981161091887} 45.1 \\
 & RA-IT & \cellcolor[rgb]{0.9937254901960785, 0.9976470588235294, 0.9921568627450981} 58.6 & \cellcolor[rgb]{0.9937254901960785, 0.9976470588235294, 0.9921568627450981} 49.1 \\
 & \ours & \cellcolor[rgb]{0.9483644752018454, 0.9794971164936562, 0.9430772779700115} 54.2 & \cellcolor[rgb]{0.9521045751633987, 0.9810718954248366, 0.946718954248366} 45.2 \\
 & \ours DPO & \cellcolor[rgb]{0.8, 0.8533333333333333, 0.8211764705882353} 43.1 & \cellcolor[rgb]{0.8, 0.8533333333333333, 0.8211764705882353} 34.5 \\
\cline{1-4}
\bottomrule
\end{tabular}

%% file: tab/retrievals.tex
\begin{tabular}{lS[table-format=2.1]S[table-format=2.1]S[table-format=2.1]S[table-format=2.1]S[table-format=2.1]}
\toprule
{Retrievals} & {0} & {1} & {2} & {4} & {8} \\
{Model} & {} & {} & {} & {} & {} \\
\midrule
\llamainstruct & \cellcolor[rgb]{0.8, 0.8674571318723567, 0.8270326797385621} 69.7 & \cellcolor[rgb]{0.9839815455594002, 0.9938577470203768, 0.9807889273356402} 70.6 & \cellcolor[rgb]{0.9511695501730104, 0.9806782006920415, 0.9458085351787774} 73.9 & \cellcolor[rgb]{0.9390142252979623, 0.9755601691657055, 0.9339730872741253} 76.1 & \cellcolor[rgb]{0.907035755478662, 0.9612179930795848, 0.9057500961168781} 78.4 \\
IT & \cellcolor[rgb]{0.9203229527104959, 0.9674186851211073, 0.9166751249519416} 55.1 & \cellcolor[rgb]{0.9937254901960785, 0.9976470588235294, 0.9921568627450981} 69.9 & \cellcolor[rgb]{0.9937254901960785, 0.9976470588235294, 0.9921568627450981} 71.8 & \cellcolor[rgb]{0.9937254901960785, 0.9976470588235294, 0.9921568627450981} 73.6 & \cellcolor[rgb]{0.9937254901960785, 0.9976470588235294, 0.9921568627450981} 75.5 \\
RA-IT & \cellcolor[rgb]{0.9937254901960785, 0.9976470588235294, 0.9921568627450981} 44.5 & \cellcolor[rgb]{0.9915109573241061, 0.9967858515955402, 0.9895732410611303} 70.0 & \cellcolor[rgb]{0.8, 0.8805720876585929, 0.8324705882352941} 78.9 & \cellcolor[rgb]{0.8259438677431756, 0.9077277970011534, 0.853041138023837} 79.4 & \cellcolor[rgb]{0.8401845444059977, 0.9226020761245675, 0.8643044982698962} 79.9 \\
\ours & \cellcolor[rgb]{0.8156093810073048, 0.8988696655132641, 0.8456593617839292} 65.9 & \cellcolor[rgb]{0.8, 0.8583775470972703, 0.8232679738562092} 78.1 & \cellcolor[rgb]{0.8, 0.8533333333333333, 0.8211764705882353} 79.8 & \cellcolor[rgb]{0.8, 0.8664482891195694, 0.8266143790849674} 80.9 & \cellcolor[rgb]{0.8, 0.8725013456362938, 0.8291241830065359} 81.4 \\
\ours DPO & \cellcolor[rgb]{0.8, 0.8533333333333333, 0.8211764705882353} 71.1 & \cellcolor[rgb]{0.8, 0.8533333333333333, 0.8211764705882353} 78.2 & \cellcolor[rgb]{0.8, 0.8553510188389081, 0.8220130718954248} 79.7 & \cellcolor[rgb]{0.8, 0.8533333333333333, 0.8211764705882353} 81.3 & \cellcolor[rgb]{0.8, 0.8533333333333333, 0.8211764705882353} 81.9 \\
\bottomrule
\end{tabular}

%% file: tab/correctness_rate.tex
\begin{tabular}{llS|SSSSSSSSS}
\toprule
{} & {} & {\textbf{Avg.}} & {PSR} & {FEVER} & {HPQA} & {MMLU} & {NQ} & {TQA} & {T-REx} & {WoW} & {zsRE} \\
\midrule
\multirow[c]{5}{*}{8B} & \llamainstruct & \cellcolor[rgb]{0.9390142252979623, 0.9755601691657055, 0.9339730872741253} 76.1 & \cellcolor[rgb]{0.9937254901960785, 0.9976470588235294, 0.9921568627450981} 67.3 & \cellcolor[rgb]{0.8387081891580161, 0.9210272971933872, 0.8631234140715109} 85.8 & \cellcolor[rgb]{0.9125720876585929, 0.9638016147635525, 0.9103021914648212} 64.0 & \cellcolor[rgb]{0.8937485582468281, 0.9550173010380623, 0.8948250672818147} 86.0 & \cellcolor[rgb]{0.8564705882352941, 0.9368089196462899, 0.8756324490580546} 70.0 & \cellcolor[rgb]{0.9937254901960785, 0.9976470588235294, 0.9921568627450981} 79.6 & \cellcolor[rgb]{0.9158938869665513, 0.9653517877739332, 0.913033448673587} 79.7 & \cellcolor[rgb]{0.9937254901960785, 0.9976470588235294, 0.9921568627450981} 64.4 & \cellcolor[rgb]{0.8539607843137255, 0.9355786236063053, 0.87440215301807} 87.9 \\
 & IT & \cellcolor[rgb]{0.9937254901960785, 0.9976470588235294, 0.9921568627450981} 73.6 & \cellcolor[rgb]{0.8715294117647059, 0.9441906958861976, 0.8830142252979623} 75.0 & \cellcolor[rgb]{0.9937254901960785, 0.9976470588235294, 0.9921568627450981} 74.7 & \cellcolor[rgb]{0.9937254901960785, 0.9976470588235294, 0.9921568627450981} 59.3 & \cellcolor[rgb]{0.9937254901960785, 0.9976470588235294, 0.9921568627450981} 81.1 & \cellcolor[rgb]{0.9937254901960785, 0.9976470588235294, 0.9921568627450981} 60.0 & \cellcolor[rgb]{0.9915109573241061, 0.9967858515955402, 0.9895732410611303} 79.7 & \cellcolor[rgb]{0.9937254901960785, 0.9976470588235294, 0.9921568627450981} 76.6 & \cellcolor[rgb]{0.8915340253748558, 0.9539838523644752, 0.8930042291426374} 72.4 & \cellcolor[rgb]{0.9937254901960785, 0.9976470588235294, 0.9921568627450981} 84.0 \\
 & RA-IT & \cellcolor[rgb]{0.8259438677431756, 0.9077277970011534, 0.853041138023837} 79.4 & \cellcolor[rgb]{0.8, 0.8533333333333333, 0.8211764705882353} 81.0 & \cellcolor[rgb]{0.8320645905420991, 0.9139407920030758, 0.8578085351787774} 86.3 & \cellcolor[rgb]{0.819915417147251, 0.902560553633218, 0.8487351018838908} 67.9 & \cellcolor[rgb]{0.9125720876585929, 0.9638016147635525, 0.9103021914648212} 85.3 & \cellcolor[rgb]{0.9305990003844675, 0.9720169165705498, 0.9257793156478278} 65.9 & \cellcolor[rgb]{0.8, 0.8533333333333333, 0.8211764705882353} 84.2 & \cellcolor[rgb]{0.8903529411764706, 0.9534179161860823, 0.892241445597847} 80.4 & \cellcolor[rgb]{0.8320645905420991, 0.9139407920030758, 0.8578085351787774} 76.0 & \cellcolor[rgb]{0.8815686274509804, 0.9491118800461361, 0.8879354094579008} 87.4 \\
 & \ours & \cellcolor[rgb]{0.8, 0.8664482891195694, 0.8266143790849674} 80.9 & \cellcolor[rgb]{0.8, 0.8846074586697424, 0.8341437908496732} 79.3 & \cellcolor[rgb]{0.8164705882352941, 0.899607843137255, 0.8462745098039215} 87.6 & \cellcolor[rgb]{0.8401845444059977, 0.9226020761245675, 0.8643044982698962} 66.8 & \cellcolor[rgb]{0.8865882352941177, 0.9515724721261054, 0.89039600153787} 86.3 & \cellcolor[rgb]{0.8069973087274125, 0.8914878892733564, 0.8395078815840061} 73.8 & \cellcolor[rgb]{0.8242214532871972, 0.9062514417531718, 0.8518108419838524} 83.1 & \cellcolor[rgb]{0.8190542099192618, 0.9018223760092272, 0.8481199538638985} 82.6 & \cellcolor[rgb]{0.8, 0.8533333333333333, 0.8211764705882353} 80.4 & \cellcolor[rgb]{0.8552156862745098, 0.9361937716262976, 0.8750173010380623} 87.9 \\
 & \ours DPO & \cellcolor[rgb]{0.8, 0.8533333333333333, 0.8211764705882353} 81.3 & \cellcolor[rgb]{0.9192156862745098, 0.9669019607843137, 0.9157647058823529} 72.8 & \cellcolor[rgb]{0.8, 0.8533333333333333, 0.8211764705882353} 90.8 & \cellcolor[rgb]{0.8, 0.8533333333333333, 0.8211764705882353} 70.3 & \cellcolor[rgb]{0.8, 0.8533333333333333, 0.8211764705882353} 91.2 & \cellcolor[rgb]{0.8, 0.8533333333333333, 0.8211764705882353} 76.4 & \cellcolor[rgb]{0.8, 0.8815809304113802, 0.8328888888888889} 83.7 & \cellcolor[rgb]{0.8, 0.8533333333333333, 0.8211764705882353} 84.3 & \cellcolor[rgb]{0.9081430219146482, 0.9617347174163783, 0.9066605151864667} 71.4 & \cellcolor[rgb]{0.8, 0.8533333333333333, 0.8211764705882353} 90.4 \\
\hhline{*{12}{=}}
\multirow[c]{5}{*}{70B} & \llamainstruct & \cellcolor[rgb]{0.980881199538639, 0.9926520569011918, 0.9771718569780854} 74.0 & \cellcolor[rgb]{0.9919538638985006, 0.996958093041138, 0.9900899653979239} 63.2 & \cellcolor[rgb]{0.8, 0.8624129181084198, 0.8249411764705883} 92.6 & \cellcolor[rgb]{0.8276355247981546, 0.9092164552095348, 0.8542652825836217} 69.5 & \cellcolor[rgb]{0.845351787773933, 0.9281138023836986, 0.8684382929642445} 84.9 & \cellcolor[rgb]{0.8401845444059977, 0.9226020761245675, 0.8643044982698962} 72.1 & \cellcolor[rgb]{0.9654901960784313, 0.9865098039215686, 0.9606274509803922} 84.2 & \cellcolor[rgb]{0.9581084198385237, 0.9835570934256055, 0.9527535563244905} 77.6 & \cellcolor[rgb]{0.9937254901960785, 0.9976470588235294, 0.9921568627450981} 35.6 & \cellcolor[rgb]{0.9418193002691273, 0.9767412533640907, 0.9367043444828912} 86.3 \\
 & IT & \cellcolor[rgb]{0.9937254901960785, 0.9976470588235294, 0.9921568627450981} 73.2 & \cellcolor[rgb]{0.9937254901960785, 0.9976470588235294, 0.9921568627450981} 63.0 & \cellcolor[rgb]{0.9937254901960785, 0.9976470588235294, 0.9921568627450981} 85.3 & \cellcolor[rgb]{0.867764705882353, 0.9423452518262206, 0.8811687812379854} 68.2 & \cellcolor[rgb]{0.9937254901960785, 0.9976470588235294, 0.9921568627450981} 79.5 & \cellcolor[rgb]{0.9305990003844675, 0.9720169165705498, 0.9257793156478278} 66.8 & \cellcolor[rgb]{0.8416608996539792, 0.9241768550557478, 0.8654855824682814} 86.6 & \cellcolor[rgb]{0.9937254901960785, 0.9976470588235294, 0.9921568627450981} 75.7 & \cellcolor[rgb]{0.8727843137254901, 0.9448058439061899, 0.8836293733179547} 48.9 & \cellcolor[rgb]{0.9937254901960785, 0.9976470588235294, 0.9921568627450981} 84.3 \\
 & RA-IT & \cellcolor[rgb]{0.9399492502883506, 0.9759538638985006, 0.934883506343714} 75.5 & \cellcolor[rgb]{0.9203229527104959, 0.9674186851211073, 0.9166751249519416} 67.9 & \cellcolor[rgb]{0.9081430219146482, 0.9617347174163783, 0.9066605151864667} 88.7 & \cellcolor[rgb]{0.8, 0.8533333333333333, 0.8211764705882353} 71.5 & \cellcolor[rgb]{0.8690196078431373, 0.9429603998462129, 0.8817839292579777} 84.2 & \cellcolor[rgb]{0.9937254901960785, 0.9976470588235294, 0.9921568627450981} 60.9 & \cellcolor[rgb]{0.8, 0.8533333333333333, 0.8211764705882353} 88.3 & \cellcolor[rgb]{0.9258592848904268, 0.9700023068050749, 0.9212272202998846} 78.6 & \cellcolor[rgb]{0.819915417147251, 0.902560553633218, 0.8487351018838908} 54.3 & \cellcolor[rgb]{0.9853102652825836, 0.9943744713571703, 0.9823391003460208} 84.8 \\
 & \ours & \cellcolor[rgb]{0.8, 0.8533333333333333, 0.8211764705882353} 80.4 & \cellcolor[rgb]{0.8, 0.8533333333333333, 0.8211764705882353} 75.5 & \cellcolor[rgb]{0.8104421376393695, 0.8944405997693194, 0.8419684736639754} 91.6 & \cellcolor[rgb]{0.8, 0.8735101883890811, 0.8295424836601307} 70.8 & \cellcolor[rgb]{0.8, 0.8533333333333333, 0.8211764705882353} 87.7 & \cellcolor[rgb]{0.8, 0.8533333333333333, 0.8211764705882353} 77.4 & \cellcolor[rgb]{0.83280276816609, 0.9147281814686659, 0.8583990772779699} 86.9 & \cellcolor[rgb]{0.8, 0.8533333333333333, 0.8211764705882353} 83.6 & \cellcolor[rgb]{0.8, 0.8533333333333333, 0.8211764705882353} 59.5 & \cellcolor[rgb]{0.8, 0.8533333333333333, 0.8211764705882353} 90.7 \\
 & \ours DPO & \cellcolor[rgb]{0.8665098039215686, 0.9417301038062283, 0.8805536332179931} 77.3 & \cellcolor[rgb]{0.8627450980392157, 0.9398846597462515, 0.8787081891580162} 70.3 & \cellcolor[rgb]{0.8, 0.8533333333333333, 0.8211764705882353} 92.9 & \cellcolor[rgb]{0.9937254901960785, 0.9976470588235294, 0.9921568627450981} 63.9 & \cellcolor[rgb]{0.8298500576701269, 0.9115786236063053, 0.8560369088811995} 85.5 & \cellcolor[rgb]{0.8044136870434448, 0.889273356401384, 0.8376624375240292} 75.0 & \cellcolor[rgb]{0.9937254901960785, 0.9976470588235294, 0.9921568627450981} 83.1 & \cellcolor[rgb]{0.8342791234140715, 0.9163029603998462, 0.8595801614763552} 81.3 & \cellcolor[rgb]{0.8044136870434448, 0.889273356401384, 0.8376624375240292} 55.9 & \cellcolor[rgb]{0.8652549019607844, 0.941114955786236, 0.8799384851980008} 88.0 \\
\cline{1-12}
\bottomrule
\end{tabular}

%% file: tab/answerable_accuracy.tex
\begin{tabular}{llS|SSSSSSSSS}
\toprule
{} & {} & {\textbf{Avg.}} & {PSR} & {FEVER} & {HPQA} & {MMLU} & {NQ} & {TQA} & {T-REx} & {WoW} & {zsRE} \\
\midrule
\multirow[c]{5}{*}{8B} & \llamainstruct & \cellcolor[rgb]{0.9937254901960785, 0.9976470588235294, 0.9921568627450981} 78.4 & \cellcolor[rgb]{0.9937254901960785, 0.9976470588235294, 0.9921568627450981} 70.9 & \cellcolor[rgb]{0.8815686274509804, 0.9491118800461361, 0.8879354094579008} 85.9 & \cellcolor[rgb]{0.8460899653979239, 0.9289011918492888, 0.8690288350634371} 79.1 & \cellcolor[rgb]{0.8409227220299884, 0.9233894655901576, 0.8648950403690888} 89.6 & \cellcolor[rgb]{0.8018300653594771, 0.8870588235294118, 0.8358169934640522} 84.4 & \cellcolor[rgb]{0.9937254901960785, 0.9976470588235294, 0.9921568627450981} 88.7 & \cellcolor[rgb]{0.9937254901960785, 0.9976470588235294, 0.9921568627450981} 70.8 & \cellcolor[rgb]{0.9937254901960785, 0.9976470588235294, 0.9921568627450981} 56.1 & \cellcolor[rgb]{0.9813241061130334, 0.9928242983467896, 0.9776885813148789} 80.6 \\
 & IT & \cellcolor[rgb]{0.9258592848904268, 0.9700023068050749, 0.9212272202998846} 79.9 & \cellcolor[rgb]{0.8283737024221454, 0.910003844675125, 0.8548558246828143} 80.5 & \cellcolor[rgb]{0.9937254901960785, 0.9976470588235294, 0.9921568627450981} 78.1 & \cellcolor[rgb]{0.8026912725874663, 0.8877970011534025, 0.8364321414840445} 81.1 & \cellcolor[rgb]{0.9937254901960785, 0.9976470588235294, 0.9921568627450981} 83.7 & \cellcolor[rgb]{0.9937254901960785, 0.9976470588235294, 0.9921568627450981} 71.8 & \cellcolor[rgb]{0.823360246059208, 0.9055132641291811, 0.8511956939638601} 90.2 & \cellcolor[rgb]{0.8, 0.8533333333333333, 0.8211764705882353} 78.9 & \cellcolor[rgb]{0.8164705882352941, 0.899607843137255, 0.8462745098039215} 69.8 & \cellcolor[rgb]{0.8, 0.8533333333333333, 0.8211764705882353} 85.4 \\
 & RA-IT & \cellcolor[rgb]{0.9103575547866205, 0.9627681660899654, 0.908481353325644} 80.2 & \cellcolor[rgb]{0.8, 0.8533333333333333, 0.8211764705882353} 83.7 & \cellcolor[rgb]{0.9014994232987312, 0.958634371395617, 0.9011980007689351} 84.9 & \cellcolor[rgb]{0.9937254901960785, 0.9976470588235294, 0.9921568627450981} 72.8 & \cellcolor[rgb]{0.9706574394463667, 0.9885767012687428, 0.9661391772395233} 85.2 & \cellcolor[rgb]{0.9640138408304498, 0.985919261822376, 0.9590526720492119} 74.9 & \cellcolor[rgb]{0.8173317954632834, 0.9003460207612457, 0.8468896578239139} 90.3 & \cellcolor[rgb]{0.892641291810842, 0.9545005767012688, 0.8939146482122261} 74.8 & \cellcolor[rgb]{0.8, 0.8533333333333333, 0.8211764705882353} 73.2 & \cellcolor[rgb]{0.940884275278739, 0.9763475586312956, 0.9357939254133025} 81.7 \\
 & \ours & \cellcolor[rgb]{0.8, 0.8563598615916955, 0.8224313725490195} 82.3 & \cellcolor[rgb]{0.8242214532871972, 0.9062514417531718, 0.8518108419838524} 80.7 & \cellcolor[rgb]{0.8156093810073048, 0.8988696655132641, 0.8456593617839292} 89.9 & \cellcolor[rgb]{0.8173317954632834, 0.9003460207612457, 0.8468896578239139} 80.5 & \cellcolor[rgb]{0.8438754325259515, 0.9265390234525183, 0.8672572087658592} 89.4 & \cellcolor[rgb]{0.8, 0.883598615916955, 0.8337254901960784} 84.7 & \cellcolor[rgb]{0.8250826605151864, 0.9069896193771626, 0.8524259900038447} 90.2 & \cellcolor[rgb]{0.9014994232987312, 0.958634371395617, 0.9011980007689351} 74.5 & \cellcolor[rgb]{0.8207766243752403, 0.9032987312572087, 0.8493502499038831} 69.4 & \cellcolor[rgb]{0.9334040753556324, 0.973198000768935, 0.9285105728565937} 81.8 \\
 & \ours DPO & \cellcolor[rgb]{0.8, 0.8533333333333333, 0.8211764705882353} 82.4 & \cellcolor[rgb]{0.8937485582468281, 0.9550173010380623, 0.8948250672818147} 77.2 & \cellcolor[rgb]{0.8, 0.8533333333333333, 0.8211764705882353} 92.8 & \cellcolor[rgb]{0.8, 0.8533333333333333, 0.8211764705882353} 82.4 & \cellcolor[rgb]{0.8, 0.8533333333333333, 0.8211764705882353} 92.4 & \cellcolor[rgb]{0.8, 0.8533333333333333, 0.8211764705882353} 86.4 & \cellcolor[rgb]{0.8, 0.8533333333333333, 0.8211764705882353} 90.7 & \cellcolor[rgb]{0.993282583621684, 0.9974748173779315, 0.9916401384083045} 70.8 & \cellcolor[rgb]{0.8276355247981546, 0.9092164552095348, 0.8542652825836217} 68.9 & \cellcolor[rgb]{0.9937254901960785, 0.9976470588235294, 0.9921568627450981} 79.9 \\
\hhline{*{12}{=}}
\multirow[c]{5}{*}{70B} & \llamainstruct & \cellcolor[rgb]{0.9492995001922337, 0.9798908112264514, 0.9439876970396002} 79.4 & \cellcolor[rgb]{0.9324690503652442, 0.97280430603614, 0.927600153787005} 73.1 & \cellcolor[rgb]{0.8, 0.8624129181084198, 0.8249411764705883} 94.4 & \cellcolor[rgb]{0.9181084198385236, 0.9663852364475202, 0.9148542868127643} 79.9 & \cellcolor[rgb]{0.9147866205305651, 0.9648350634371395, 0.9121230296039985} 85.9 & \cellcolor[rgb]{0.8, 0.8694748173779315, 0.8278692810457516} 86.7 & \cellcolor[rgb]{0.986638985005767, 0.9948911956939639, 0.9838892733564014} 90.4 & \cellcolor[rgb]{0.899284890426759, 0.95760092272203, 0.8993771626297578} 77.9 & \cellcolor[rgb]{0.9937254901960785, 0.9976470588235294, 0.9921568627450981} 39.3 & \cellcolor[rgb]{0.8514509803921568, 0.9343483275663207, 0.8731718569780853} 87.0 \\
 & IT & \cellcolor[rgb]{0.9937254901960785, 0.9976470588235294, 0.9921568627450981} 77.3 & \cellcolor[rgb]{0.9937254901960785, 0.9976470588235294, 0.9921568627450981} 67.8 & \cellcolor[rgb]{0.9937254901960785, 0.9976470588235294, 0.9921568627450981} 74.3 & \cellcolor[rgb]{0.8, 0.8533333333333333, 0.8211764705882353} 82.4 & \cellcolor[rgb]{0.9937254901960785, 0.9976470588235294, 0.9921568627450981} 83.5 & \cellcolor[rgb]{0.8665098039215686, 0.9417301038062283, 0.8805536332179931} 78.7 & \cellcolor[rgb]{0.8778039215686274, 0.9472664359861591, 0.8860899653979238} 92.3 & \cellcolor[rgb]{0.8752941176470588, 0.9460361399461745, 0.8848596693579392} 78.9 & \cellcolor[rgb]{0.8878431372549019, 0.9521876201460977, 0.8910111495578623} 51.6 & \cellcolor[rgb]{0.884078431372549, 0.9503421760861207, 0.8891657054978854} 85.9 \\
 & RA-IT & \cellcolor[rgb]{0.9901822376009227, 0.9962691272587466, 0.9880230680507497} 77.5 & \cellcolor[rgb]{0.9835386389850058, 0.993685505574779, 0.9802722029988467} 69.2 & \cellcolor[rgb]{0.8790588235294118, 0.9478815840061515, 0.8867051134179162} 85.5 & \cellcolor[rgb]{0.8765490196078431, 0.9466512879661668, 0.8854748173779315} 80.5 & \cellcolor[rgb]{0.9799953863898501, 0.9923075740099961, 0.9761384083044983} 84.2 & \cellcolor[rgb]{0.9937254901960785, 0.9976470588235294, 0.9921568627450981} 65.9 & \cellcolor[rgb]{0.8, 0.8533333333333333, 0.8211764705882353} 94.1 & \cellcolor[rgb]{0.9081430219146482, 0.9617347174163783, 0.9066605151864667} 77.5 & \cellcolor[rgb]{0.8283737024221454, 0.910003844675125, 0.8548558246828143} 57.3 & \cellcolor[rgb]{0.9455594002306805, 0.9783160322952711, 0.9403460207612456} 83.5 \\
 & \ours & \cellcolor[rgb]{0.955844675124952, 0.9826466743560169, 0.9503606305267205} 79.2 & \cellcolor[rgb]{0.8, 0.8664482891195694, 0.8266143790849674} 82.1 & \cellcolor[rgb]{0.9427543252595155, 0.9771349480968858, 0.9376147635524799} 80.7 & \cellcolor[rgb]{0.9691810841983852, 0.9879861591695501, 0.964564398308343} 79.0 & \cellcolor[rgb]{0.9928396770472895, 0.9973025759323337, 0.9911234140715109} 83.5 & \cellcolor[rgb]{0.8113033448673587, 0.8951787773933102, 0.8425836216839677} 84.2 & \cellcolor[rgb]{0.9937254901960785, 0.9976470588235294, 0.9921568627450981} 90.1 & \cellcolor[rgb]{0.9937254901960785, 0.9976470588235294, 0.9921568627450981} 71.9 & \cellcolor[rgb]{0.8, 0.8825897731641676, 0.8333071895424836} 60.6 & \cellcolor[rgb]{0.9937254901960785, 0.9976470588235294, 0.9921568627450981} 80.5 \\
 & \ours DPO & \cellcolor[rgb]{0.8, 0.8533333333333333, 0.8211764705882353} 84.8 & \cellcolor[rgb]{0.8, 0.8533333333333333, 0.8211764705882353} 82.8 & \cellcolor[rgb]{0.8, 0.8533333333333333, 0.8211764705882353} 95.1 & \cellcolor[rgb]{0.9937254901960785, 0.9976470588235294, 0.9921568627450981} 78.2 & \cellcolor[rgb]{0.8, 0.8533333333333333, 0.8211764705882353} 89.2 & \cellcolor[rgb]{0.8, 0.8533333333333333, 0.8211764705882353} 88.1 & \cellcolor[rgb]{0.9549096501345636, 0.9822529796232218, 0.9494502114571318} 91.2 & \cellcolor[rgb]{0.8, 0.8533333333333333, 0.8211764705882353} 84.6 & \cellcolor[rgb]{0.8, 0.8533333333333333, 0.8211764705882353} 63.4 & \cellcolor[rgb]{0.8, 0.8533333333333333, 0.8211764705882353} 90.9 \\
\cline{1-12}
\bottomrule
\end{tabular}

%% file: tab/f1.tex
\begin{tabular}{llS|SSSSSSSSS}
\toprule
{} & {} & {\textbf{Avg.}} & {PSR} & {FEVER} & {HPQA} & {MMLU} & {NQ} & {TQA} & {T-REx} & {WoW} & {zsRE} \\
\midrule
\multirow[c]{5}{*}{8B} & \llamainstruct & \cellcolor[rgb]{0.9826528258362168, 0.9933410226835833, 0.9792387543252595} 76.9 & \cellcolor[rgb]{0.9937254901960785, 0.9976470588235294, 0.9921568627450981} 68.0 & \cellcolor[rgb]{0.8539607843137255, 0.9355786236063053, 0.87440215301807} 85.8 & \cellcolor[rgb]{0.9446243752402922, 0.9779223375624759, 0.9394356016916571} 70.7 & \cellcolor[rgb]{0.8665098039215686, 0.9417301038062283, 0.8805536332179931} 87.8 & \cellcolor[rgb]{0.8350173010380623, 0.9170903498654364, 0.8601707035755478} 76.5 & \cellcolor[rgb]{0.9937254901960785, 0.9976470588235294, 0.9921568627450981} 83.9 & \cellcolor[rgb]{0.9937254901960785, 0.9976470588235294, 0.9921568627450981} 75.0 & \cellcolor[rgb]{0.9937254901960785, 0.9976470588235294, 0.9921568627450981} 60.0 & \cellcolor[rgb]{0.9937254901960785, 0.9976470588235294, 0.9921568627450981} 84.1 \\
 & IT & \cellcolor[rgb]{0.9937254901960785, 0.9976470588235294, 0.9921568627450981} 76.4 & \cellcolor[rgb]{0.8475663206459054, 0.9304759707804691, 0.8702099192618223} 76.5 & \cellcolor[rgb]{0.9937254901960785, 0.9976470588235294, 0.9921568627450981} 76.3 & \cellcolor[rgb]{0.9937254901960785, 0.9976470588235294, 0.9921568627450981} 68.5 & \cellcolor[rgb]{0.9937254901960785, 0.9976470588235294, 0.9921568627450981} 82.3 & \cellcolor[rgb]{0.9937254901960785, 0.9976470588235294, 0.9921568627450981} 65.4 & \cellcolor[rgb]{0.9610611303344867, 0.9847381776239907, 0.9559031141868513} 84.6 & \cellcolor[rgb]{0.8147481737793156, 0.8981314878892733, 0.8450442137639369} 77.7 & \cellcolor[rgb]{0.8250826605151864, 0.9069896193771626, 0.8524259900038447} 71.1 & \cellcolor[rgb]{0.8216378316032296, 0.9040369088811995, 0.8499653979238755} 84.7 \\
 & RA-IT & \cellcolor[rgb]{0.8514509803921568, 0.9343483275663207, 0.8731718569780853} 79.6 & \cellcolor[rgb]{0.8, 0.8533333333333333, 0.8211764705882353} 81.3 & \cellcolor[rgb]{0.8589803921568627, 0.9380392156862745, 0.8768627450980392} 85.6 & \cellcolor[rgb]{0.9588465974625144, 0.9838523644752019, 0.9535409457900808} 70.3 & \cellcolor[rgb]{0.9418193002691273, 0.9767412533640907, 0.9367043444828912} 85.3 & \cellcolor[rgb]{0.9446243752402922, 0.9779223375624759, 0.9394356016916571} 70.1 & \cellcolor[rgb]{0.8, 0.8533333333333333, 0.8211764705882353} 87.1 & \cellcolor[rgb]{0.8283737024221454, 0.910003844675125, 0.8548558246828143} 77.5 & \cellcolor[rgb]{0.8, 0.8533333333333333, 0.8211764705882353} 74.6 & \cellcolor[rgb]{0.9081430219146482, 0.9617347174163783, 0.9066605151864667} 84.4 \\
 & \ours & \cellcolor[rgb]{0.8, 0.8644306036139946, 0.8257777777777777} 81.3 & \cellcolor[rgb]{0.8113033448673587, 0.8951787773933102, 0.8425836216839677} 78.9 & \cellcolor[rgb]{0.8156093810073048, 0.8988696655132641, 0.8456593617839292} 88.7 & \cellcolor[rgb]{0.8564705882352941, 0.9368089196462899, 0.8756324490580546} 73.0 & \cellcolor[rgb]{0.8665098039215686, 0.9417301038062283, 0.8805536332179931} 87.8 & \cellcolor[rgb]{0.8035524798154556, 0.8885351787773933, 0.8370472895040368} 78.9 & \cellcolor[rgb]{0.8138869665513264, 0.8973933102652826, 0.8444290657439446} 86.5 & \cellcolor[rgb]{0.8, 0.8533333333333333, 0.8211764705882353} 78.4 & \cellcolor[rgb]{0.8, 0.8543421760861207, 0.8215947712418301} 74.5 & \cellcolor[rgb]{0.8, 0.883598615916955, 0.8337254901960784} 84.8 \\
 & \ours DPO & \cellcolor[rgb]{0.8, 0.8533333333333333, 0.8211764705882353} 81.5 & \cellcolor[rgb]{0.9048212226066897, 0.9601845444059977, 0.9039292579777009} 74.0 & \cellcolor[rgb]{0.8, 0.8533333333333333, 0.8211764705882353} 91.8 & \cellcolor[rgb]{0.8, 0.8533333333333333, 0.8211764705882353} 75.9 & \cellcolor[rgb]{0.8, 0.8533333333333333, 0.8211764705882353} 91.8 & \cellcolor[rgb]{0.8, 0.8533333333333333, 0.8211764705882353} 81.1 & \cellcolor[rgb]{0.8, 0.8593863898500577, 0.8236862745098039} 87.1 & \cellcolor[rgb]{0.8652549019607844, 0.941114955786236, 0.8799384851980008} 77.0 & \cellcolor[rgb]{0.8372318339100346, 0.9194525182622069, 0.8619423298731257} 70.2 & \cellcolor[rgb]{0.8, 0.8533333333333333, 0.8211764705882353} 84.9 \\
\hhline{*{12}{=}}
\multirow[c]{5}{*}{70B} & \llamainstruct & \cellcolor[rgb]{0.9539746251441753, 0.9818592848904267, 0.9485397923875433} 76.4 & \cellcolor[rgb]{0.9684429065743945, 0.9876908881199539, 0.9637770088427527} 66.9 & \cellcolor[rgb]{0.8, 0.8634217608612073, 0.825359477124183} 93.4 & \cellcolor[rgb]{0.826805074971165, 0.9084659746251441, 0.8536562860438293} 74.3 & \cellcolor[rgb]{0.8438754325259515, 0.9265390234525183, 0.8672572087658592} 85.3 & \cellcolor[rgb]{0.8009688581314879, 0.8863206459054209, 0.83520184544406} 78.7 & \cellcolor[rgb]{0.9879677047289505, 0.9954079200307574, 0.985439446366782} 87.2 & \cellcolor[rgb]{0.9844244521337947, 0.9940299884659747, 0.9813056516724337} 77.7 & \cellcolor[rgb]{0.9937254901960785, 0.9976470588235294, 0.9921568627450981} 37.4 & \cellcolor[rgb]{0.9003921568627451, 0.9581176470588235, 0.9002875816993464} 86.6 \\
 & IT & \cellcolor[rgb]{0.9937254901960785, 0.9976470588235294, 0.9921568627450981} 74.9 & \cellcolor[rgb]{0.9937254901960785, 0.9976470588235294, 0.9921568627450981} 64.5 & \cellcolor[rgb]{0.9937254901960785, 0.9976470588235294, 0.9921568627450981} 79.3 & \cellcolor[rgb]{0.8156093810073048, 0.8988696655132641, 0.8456593617839292} 74.6 & \cellcolor[rgb]{0.9937254901960785, 0.9976470588235294, 0.9921568627450981} 81.3 & \cellcolor[rgb]{0.8890980392156863, 0.95280276816609, 0.8916262975778546} 72.2 & \cellcolor[rgb]{0.8665098039215686, 0.9417301038062283, 0.8805536332179931} 89.4 & \cellcolor[rgb]{0.9937254901960785, 0.9976470588235294, 0.9921568627450981} 77.2 & \cellcolor[rgb]{0.8702745098039215, 0.9435755478662053, 0.88239907727797} 50.2 & \cellcolor[rgb]{0.969919261822376, 0.9882814302191465, 0.9653517877739332} 85.1 \\
 & RA-IT & \cellcolor[rgb]{0.9581084198385237, 0.9835570934256055, 0.9527535563244905} 76.3 & \cellcolor[rgb]{0.9581084198385237, 0.9835570934256055, 0.9527535563244905} 67.6 & \cellcolor[rgb]{0.8865882352941177, 0.9515724721261054, 0.89039600153787} 86.9 & \cellcolor[rgb]{0.8, 0.8533333333333333, 0.8211764705882353} 75.7 & \cellcolor[rgb]{0.9048212226066897, 0.9601845444059977, 0.9039292579777009} 84.0 & \cellcolor[rgb]{0.9937254901960785, 0.9976470588235294, 0.9921568627450981} 63.3 & \cellcolor[rgb]{0.8, 0.8533333333333333, 0.8211764705882353} 91.1 & \cellcolor[rgb]{0.976562860438293, 0.9909388696655133, 0.9724382929642446} 78.1 & \cellcolor[rgb]{0.8138869665513264, 0.8973933102652826, 0.8444290657439446} 55.7 & \cellcolor[rgb]{0.9937254901960785, 0.9976470588235294, 0.9921568627450981} 84.1 \\
 & \ours & \cellcolor[rgb]{0.8207766243752403, 0.9032987312572087, 0.8493502499038831} 79.4 & \cellcolor[rgb]{0.8, 0.8533333333333333, 0.8211764705882353} 77.6 & \cellcolor[rgb]{0.9114648212226066, 0.9632848904267589, 0.9093917723952326} 85.6 & \cellcolor[rgb]{0.8156093810073048, 0.8988696655132641, 0.8456593617839292} 74.6 & \cellcolor[rgb]{0.8409227220299884, 0.9233894655901576, 0.8648950403690888} 85.4 & \cellcolor[rgb]{0.8, 0.8583775470972703, 0.8232679738562092} 80.6 & \cellcolor[rgb]{0.9287289504036909, 0.9712295271049596, 0.9239584775086505} 88.5 & \cellcolor[rgb]{0.993282583621684, 0.9974748173779315, 0.9916401384083045} 77.3 & \cellcolor[rgb]{0.8, 0.8533333333333333, 0.8211764705882353} 60.0 & \cellcolor[rgb]{0.9625374855824683, 0.9853287197231834, 0.9574778931180316} 85.3 \\
 & \ours DPO & \cellcolor[rgb]{0.8, 0.8533333333333333, 0.8211764705882353} 80.7 & \cellcolor[rgb]{0.8138869665513264, 0.8973933102652826, 0.8444290657439446} 75.1 & \cellcolor[rgb]{0.8, 0.8533333333333333, 0.8211764705882353} 94.0 & \cellcolor[rgb]{0.9937254901960785, 0.9976470588235294, 0.9921568627450981} 70.3 & \cellcolor[rgb]{0.8, 0.8533333333333333, 0.8211764705882353} 87.3 & \cellcolor[rgb]{0.8, 0.8533333333333333, 0.8211764705882353} 81.0 & \cellcolor[rgb]{0.9937254901960785, 0.9976470588235294, 0.9921568627450981} 87.0 & \cellcolor[rgb]{0.8, 0.8533333333333333, 0.8211764705882353} 82.9 & \cellcolor[rgb]{0.8, 0.8593863898500577, 0.8236862745098039} 59.4 & \cellcolor[rgb]{0.8, 0.8533333333333333, 0.8211764705882353} 89.4 \\
\cline{1-12}
\bottomrule
\end{tabular}

%% file: tab/false_refusals_with_baseline.tex
\begin{tabular}{llS|SSSSSSSSS}
\toprule
{} & {} & {\textbf{Avg.}} & {PSR} & {FEVER} & {HPQA} & {MMLU} & {NQ} & {TQA} & {T-REx} & {WoW} & {zsRE} \\
\midrule
\multirow[c]{3}{*}{8B} & RA-IT & \cellcolor[rgb]{0.9937254901960785, 0.9976470588235294, 0.9921568627450981} 43.1 & \cellcolor[rgb]{0.976562860438293, 0.9909388696655133, 0.9724382929642446} 15.8 & \cellcolor[rgb]{0.9937254901960785, 0.9976470588235294, 0.9921568627450981} 37.2 & \cellcolor[rgb]{0.9937254901960785, 0.9976470588235294, 0.9921568627450981} 29.0 & \cellcolor[rgb]{0.9937254901960785, 0.9976470588235294, 0.9921568627450981} 24.2 & \cellcolor[rgb]{0.9937254901960785, 0.9976470588235294, 0.9921568627450981} 56.9 & \cellcolor[rgb]{0.9937254901960785, 0.9976470588235294, 0.9921568627450981} 29.1 & \cellcolor[rgb]{0.9937254901960785, 0.9976470588235294, 0.9921568627450981} 78.7 & \cellcolor[rgb]{0.9937254901960785, 0.9976470588235294, 0.9921568627450981} 34.3 & \cellcolor[rgb]{0.9937254901960785, 0.9976470588235294, 0.9921568627450981} 82.7 \\
 & \ours & \cellcolor[rgb]{0.8350173010380623, 0.9170903498654364, 0.8601707035755478} 35.4 & \cellcolor[rgb]{0.9937254901960785, 0.9976470588235294, 0.9921568627450981} 17.4 & \cellcolor[rgb]{0.8, 0.8533333333333333, 0.8211764705882353} 26.6 & \cellcolor[rgb]{0.8379700115340254, 0.920239907727797, 0.8625328719723183} 21.1 & \cellcolor[rgb]{0.864, 0.9404998077662438, 0.8793233371780085} 15.6 & \cellcolor[rgb]{0.808719723183391, 0.8929642445213379, 0.8407381776239907} 31.0 & \cellcolor[rgb]{0.8690196078431373, 0.942960399846213, 0.8817839292579777} 25.0 & \cellcolor[rgb]{0.8, 0.8533333333333333, 0.8211764705882353} 76.3 & \cellcolor[rgb]{0.8394463667820069, 0.9218146866589773, 0.8637139561707036} 29.5 & \cellcolor[rgb]{0.8181930026912726, 0.9010841983852365, 0.8475048058439062} 76.4 \\
 & \ours DPO & \cellcolor[rgb]{0.8, 0.8533333333333333, 0.8211764705882353} 32.3 & \cellcolor[rgb]{0.8, 0.8533333333333333, 0.8211764705882353} 6.1 & \cellcolor[rgb]{0.8431372549019608, 0.9257516339869281, 0.8666666666666667} 30.1 & \cellcolor[rgb]{0.8, 0.8533333333333333, 0.8211764705882353} 17.6 & \cellcolor[rgb]{0.8, 0.8533333333333333, 0.8211764705882353} 9.4 & \cellcolor[rgb]{0.8, 0.8533333333333333, 0.8211764705882353} 25.8 & \cellcolor[rgb]{0.8, 0.8533333333333333, 0.8211764705882353} 21.9 & \cellcolor[rgb]{0.8690196078431373, 0.942960399846213, 0.8817839292579777} 77.3 & \cellcolor[rgb]{0.8, 0.8533333333333333, 0.8211764705882353} 27.3 & \cellcolor[rgb]{0.8, 0.8533333333333333, 0.8211764705882353} 74.7 \\
\hhline{*{12}{=}}
\multirow[c]{3}{*}{70B} & RA-IT & \cellcolor[rgb]{0.9937254901960785, 0.9976470588235294, 0.9921568627450981} 49.1 & \cellcolor[rgb]{0.9937254901960785, 0.9976470588235294, 0.9921568627450981} 39.1 & \cellcolor[rgb]{0.9617993079584775, 0.9850334486735871, 0.9566905036524413} 70.8 & \cellcolor[rgb]{0.845351787773933, 0.9281138023836986, 0.8684382929642445} 26.7 & \cellcolor[rgb]{0.9937254901960785, 0.9976470588235294, 0.9921568627450981} 20.1 & \cellcolor[rgb]{0.9937254901960785, 0.9976470588235294, 0.9921568627450981} 71.6 & \cellcolor[rgb]{0.826805074971165, 0.9084659746251441, 0.8536562860438293} 27.3 & \cellcolor[rgb]{0.9464944252210689, 0.9787097270280661, 0.9412564398308343} 74.8 & \cellcolor[rgb]{0.9937254901960785, 0.9976470588235294, 0.9921568627450981} 35.5 & \cellcolor[rgb]{0.9937254901960785, 0.9976470588235294, 0.9921568627450981} 76.4 \\
 & \ours & \cellcolor[rgb]{0.9521045751633987, 0.9810718954248366, 0.946718954248366} 45.2 & \cellcolor[rgb]{0.8, 0.8533333333333333, 0.8211764705882353} 21.7 & \cellcolor[rgb]{0.9937254901960785, 0.9976470588235294, 0.9921568627450981} 76.3 & \cellcolor[rgb]{0.9937254901960785, 0.9976470588235294, 0.9921568627450981} 29.2 & \cellcolor[rgb]{0.8, 0.8805720876585929, 0.8324705882352941} 17.4 & \cellcolor[rgb]{0.83280276816609, 0.9147281814686659, 0.85839907727797} 36.2 & \cellcolor[rgb]{0.9937254901960785, 0.9976470588235294, 0.9921568627450981} 39.1 & \cellcolor[rgb]{0.9937254901960785, 0.9976470588235294, 0.9921568627450981} 78.4 & \cellcolor[rgb]{0.9647520184544406, 0.9862145328719724, 0.959840061514802} 33.6 & \cellcolor[rgb]{0.9857531718569781, 0.9945467128027682, 0.9828558246828143} 74.6 \\
 & \ours DPO & \cellcolor[rgb]{0.8, 0.8533333333333333, 0.8211764705882353} 34.5 & \cellcolor[rgb]{0.8138869665513264, 0.8973933102652826, 0.8444290657439446} 25.0 & \cellcolor[rgb]{0.8, 0.8533333333333333, 0.8211764705882353} 51.5 & \cellcolor[rgb]{0.8, 0.8533333333333333, 0.8211764705882353} 25.4 & \cellcolor[rgb]{0.8, 0.8533333333333333, 0.8211764705882353} 17.1 & \cellcolor[rgb]{0.8, 0.8533333333333333, 0.8211764705882353} 22.7 & \cellcolor[rgb]{0.8, 0.8533333333333333, 0.8211764705882353} 23.4 & \cellcolor[rgb]{0.8, 0.8533333333333333, 0.8211764705882353} 66.1 & \cellcolor[rgb]{0.8, 0.8533333333333333, 0.8211764705882353} 26.4 & \cellcolor[rgb]{0.8, 0.8533333333333333, 0.8211764705882353} 52.7 \\
\cline{1-12}
\bottomrule
\end{tabular}

%% file: tab/counterfactual.tex
\begin{tabular}{llS|SSSSSSSSS}
\toprule
{} & {} & {\textbf{Avg.}} & {PSR} & {FEVER} & {HPQA} & {MMLU} & {NQ} & {TQA} & {T-REx} & {WoW} & {zsRE} \\
\midrule
\multirow[c]{4}{*}{8B} & IT & \cellcolor[rgb]{0.9937254901960785, 0.9976470588235294, 0.9921568627450981} 61.1 & \cellcolor[rgb]{0.9937254901960785, 0.9976470588235294, 0.9921568627450981} 71.4 & \cellcolor[rgb]{0.960322952710496, 0.9844429065743945, 0.955115724721261} 75.6 & \cellcolor[rgb]{0.8652549019607843, 0.941114955786236, 0.8799384851980008} 32.6 & \cellcolor[rgb]{0.9937254901960785, 0.9976470588235294, 0.9921568627450981} 81.6 & \cellcolor[rgb]{0.9937254901960785, 0.9976470588235294, 0.9921568627450981} 62.4 & \cellcolor[rgb]{0.9573702422145328, 0.9832618223760092, 0.9519661668589005} 31.0 & \cellcolor[rgb]{0.9937254901960785, 0.9976470588235294, 0.9921568627450981} 71.6 & \cellcolor[rgb]{0.9937254901960785, 0.9976470588235294, 0.9921568627450981} 52.5 & \cellcolor[rgb]{0.8627450980392157, 0.9398846597462515, 0.8787081891580162} 71.7 \\
 & RA-IT & \cellcolor[rgb]{0.9736101499423299, 0.989757785467128, 0.9692887351018838} 59.3 & \cellcolor[rgb]{0.8828235294117647, 0.9497270280661284, 0.8885505574778931} 55.6 & \cellcolor[rgb]{0.9937254901960785, 0.9976470588235294, 0.9921568627450981} 77.7 & \cellcolor[rgb]{0.9937254901960785, 0.9976470588235294, 0.9921568627450981} 39.4 & \cellcolor[rgb]{0.8915340253748558, 0.9539838523644752, 0.8930042291426374} 78.9 & \cellcolor[rgb]{0.9888535178777393, 0.9957524029219531, 0.9864728950403691} 60.6 & \cellcolor[rgb]{0.9937254901960785, 0.9976470588235294, 0.9921568627450981} 32.5 & \cellcolor[rgb]{0.892641291810842, 0.9545005767012688, 0.8939146482122261} 67.5 & \cellcolor[rgb]{0.8, 0.8533333333333333, 0.8211764705882353} 44.4 & \cellcolor[rgb]{0.9937254901960785, 0.9976470588235294, 0.9921568627450981} 77.5 \\
 & \ours & \cellcolor[rgb]{0.8035524798154556, 0.8885351787773933, 0.8370472895040368} 51.4 & \cellcolor[rgb]{0.826805074971165, 0.9084659746251441, 0.8536562860438293} 48.6 & \cellcolor[rgb]{0.8727843137254901, 0.9448058439061899, 0.8836293733179547} 72.6 & \cellcolor[rgb]{0.8, 0.8533333333333333, 0.8211764705882353} 27.7 & \cellcolor[rgb]{0.8, 0.8714925028835063, 0.8287058823529412} 76.6 & \cellcolor[rgb]{0.8035524798154556, 0.8885351787773933, 0.8370472895040368} 30.0 & \cellcolor[rgb]{0.8164705882352941, 0.899607843137255, 0.8462745098039215} 27.4 & \cellcolor[rgb]{0.8, 0.8533333333333333, 0.8211764705882353} 63.3 & \cellcolor[rgb]{0.899284890426759, 0.95760092272203, 0.8993771626297578} 48.7 & \cellcolor[rgb]{0.8, 0.8533333333333333, 0.8211764705882353} 67.6 \\
 & \ours DPO & \cellcolor[rgb]{0.8, 0.8533333333333333, 0.8211764705882353} 49.8 & \cellcolor[rgb]{0.8, 0.8533333333333333, 0.8211764705882353} 41.2 & \cellcolor[rgb]{0.8, 0.8533333333333333, 0.8211764705882353} 68.5 & \cellcolor[rgb]{0.8, 0.8846074586697423, 0.8341437908496732} 29.1 & \cellcolor[rgb]{0.8, 0.8533333333333333, 0.8211764705882353} 76.2 & \cellcolor[rgb]{0.8, 0.8533333333333333, 0.8211764705882353} 24.6 & \cellcolor[rgb]{0.8, 0.8533333333333333, 0.8211764705882353} 26.1 & \cellcolor[rgb]{0.8, 0.8583775470972703, 0.8232679738562092} 63.5 & \cellcolor[rgb]{0.9324690503652441, 0.97280430603614, 0.927600153787005} 49.7 & \cellcolor[rgb]{0.8044136870434448, 0.889273356401384, 0.8376624375240292} 69.0 \\
\hhline{*{12}{=}}
\multirow[c]{4}{*}{70B} & IT & \cellcolor[rgb]{0.9736101499423299, 0.989757785467128, 0.9692887351018838} 56.1 & \cellcolor[rgb]{0.9848673587081892, 0.9942022299115725, 0.9818223760092272} 56.8 & \cellcolor[rgb]{0.9937254901960785, 0.9976470588235294, 0.9921568627450981} 81.8 & \cellcolor[rgb]{0.8, 0.8533333333333333, 0.8211764705882353} 26.3 & \cellcolor[rgb]{0.9937254901960785, 0.9976470588235294, 0.9921568627450981} 83.2 & \cellcolor[rgb]{0.9399492502883506, 0.9759538638985006, 0.934883506343714} 54.0 & \cellcolor[rgb]{0.8009688581314879, 0.8863206459054209, 0.83520184544406} 22.6 & \cellcolor[rgb]{0.9937254901960785, 0.9976470588235294, 0.9921568627450981} 61.8 & \cellcolor[rgb]{0.8, 0.8533333333333333, 0.8211764705882353} 49.7 & \cellcolor[rgb]{0.9758246828143022, 0.990643598615917, 0.9716509034986544} 69.0 \\
 & RA-IT & \cellcolor[rgb]{0.9937254901960785, 0.9976470588235294, 0.9921568627450981} 58.6 & \cellcolor[rgb]{0.9937254901960785, 0.9976470588235294, 0.9921568627450981} 58.8 & \cellcolor[rgb]{0.8614901960784314, 0.9392695117262592, 0.8780930411380239} 72.4 & \cellcolor[rgb]{0.8702745098039215, 0.9435755478662053, 0.88239907727797} 31.3 & \cellcolor[rgb]{0.9937254901960785, 0.9976470588235294, 0.9921568627450981} 83.2 & \cellcolor[rgb]{0.9937254901960785, 0.9976470588235294, 0.9921568627450981} 69.8 & \cellcolor[rgb]{0.8, 0.8533333333333333, 0.8211764705882353} 20.9 & \cellcolor[rgb]{0.9919538638985006, 0.996958093041138, 0.9900899653979239} 61.4 & \cellcolor[rgb]{0.9780392156862745, 0.9915294117647059, 0.9740130718954249} 55.0 & \cellcolor[rgb]{0.9937254901960785, 0.9976470588235294, 0.9921568627450981} 74.1 \\
 & \ours & \cellcolor[rgb]{0.9483644752018454, 0.9794971164936562, 0.9430772779700115} 54.2 & \cellcolor[rgb]{0.8552156862745098, 0.9361937716262976, 0.8750173010380623} 43.2 & \cellcolor[rgb]{0.9026066897347174, 0.9591510957324106, 0.9021084198385236} 74.5 & \cellcolor[rgb]{0.9937254901960785, 0.9976470588235294, 0.9921568627450981} 37.9 & \cellcolor[rgb]{0.8, 0.865439446366782, 0.8261960784313725} 79.7 & \cellcolor[rgb]{0.8298500576701269, 0.9115786236063053, 0.8560369088811995} 33.2 & \cellcolor[rgb]{0.9937254901960785, 0.9976470588235294, 0.9921568627450981} 33.6 & \cellcolor[rgb]{0.9892964244521338, 0.9959246443675509, 0.9869896193771627} 61.0 & \cellcolor[rgb]{0.9937254901960785, 0.9976470588235294, 0.9921568627450981} 55.8 & \cellcolor[rgb]{0.9728719723183391, 0.9894625144175317, 0.9685013456362938} 68.4 \\
 & \ours DPO & \cellcolor[rgb]{0.8, 0.8533333333333333, 0.8211764705882353} 43.1 & \cellcolor[rgb]{0.8, 0.8533333333333333, 0.8211764705882353} 33.3 & \cellcolor[rgb]{0.8, 0.8533333333333333, 0.8211764705882353} 65.9 & \cellcolor[rgb]{0.9225374855824683, 0.9684521337946943, 0.9184959630911188} 33.4 & \cellcolor[rgb]{0.8, 0.8533333333333333, 0.8211764705882353} 79.5 & \cellcolor[rgb]{0.8, 0.8533333333333333, 0.8211764705882353} 20.1 & \cellcolor[rgb]{0.8026912725874663, 0.8877970011534025, 0.8364321414840445} 22.7 & \cellcolor[rgb]{0.8, 0.8533333333333333, 0.8211764705882353} 42.1 & \cellcolor[rgb]{0.8438754325259515, 0.9265390234525183, 0.8672572087658592} 51.8 & \cellcolor[rgb]{0.8, 0.8533333333333333, 0.8211764705882353} 38.9 \\
\cline{1-12}
\bottomrule
\end{tabular}